\documentclass{article}

\usepackage[preprint,nonatbib]{neurips_2023}

\usepackage[utf8]{inputenc} 
\usepackage[T1]{fontenc}    
\usepackage{hyperref}       
\usepackage{url}            
\usepackage{booktabs}       
\usepackage{amsfonts}       
\usepackage{nicefrac}       
\usepackage{microtype}      
\usepackage{xcolor}         
\usepackage{import}
\usepackage{adjustbox}
\usepackage{svg}
\usepackage{tikz}
\usetikzlibrary{fit,positioning, calc, decorations.pathreplacing}
\usepackage{listofitems} 
\usetikzlibrary{decorations.pathreplacing}
\usetikzlibrary{fadings}

\let\ENDGRAPH\endinput
\usepackage{float}
\usepackage{subcaption}
\usepackage{booktabs}
\usepackage{amsmath}
\usepackage{multirow}
\usepackage{enumitem}
\usepackage{siunitx}

\title{AnoRand: A Semi Supervised Deep Learning Anomaly Detection Method by Random Labeling}

\author{%
  Mansour ~Zoubeirou A Mayaki \\
 Université Côte d’Azur \\ Inria, CNRS, Nice France \\
  \texttt{zammaya76@gmail.com} \\
  \And
  Michel ~Riveill \\
  Université Côte d’Azur \\ CNRS, Inria, Nice France \\
\texttt{michel.riveill@unice.fr}
}

\begin{document}

\maketitle

\begin{abstract}
Anomaly detection or more generally outliers detection  is one of the most popular and challenging subject in theoretical and applied machine learning. The main challenge is that in general we have access to very few labeled data or no labels at all. 
In this paper, we present a new semi-supervised anomaly detection method called \textbf{AnoRand} by combining a deep learning architecture with random synthetic label generation.
The proposed architecture has two building blocks: (1) a noise detection (ND) block composed of feed forward ferceptron and (2) an autoencoder (AE) block. 
The main idea of this new architecture is to learn one class (e.g. the majority class in case of anomaly detection) as well as possible by taking advantage of the ability of auto encoders to represent data in a latent space and the ability of Feed Forward Perceptron (FFP) to learn one class when the data is highly imbalanced. First, we create synthetic anomalies by randomly disturbing (add noise) few samples (e.g. 2\%)  from the training set. Second, we use the normal and the synthetic samples as input to our model.
We compared the performance of the proposed method to 17 state-of-the-art unsupervised anomaly detection method on synthetic datasets and 57 real-world datasets. 
Our results show that this new method generally outperforms most of the state-of-the-art methods and has the best performance (AUC ROC and AUC PR) on the vast majority of reference datasets. We also tested our method in a supervised way by using the actual labels to train the model. The results show that it has very good performance compared to most of state-of-the-art supervised algorithms.
\end{abstract}

\section{Introduction}
Anomaly detection  is one of the most exciting and challenging subject in theoretical and applied machine learning. nowadays, deep learning methods are increasingly used for anomaly detection. These methods can be categorized in two main families: supervised methods and unsupervised methods. Supervised anomaly detected methods are used  when we have access to labeled data and unsupervised or semi supervised methods when there no labeled data or very few of them. 
One of the main challenge in anomaly detection or more generally  outlier detection  is that you don't have enough samples labeled as anomalous. 
In this situation, most of the classical machine learning methods fail to learn the minority (anomaly) class that most of the time the class of interest. Another challenge is that the labels are often not accurate. Some samples labeled anomalous may not be actual ones and and vice versa. Unsupervised methods have also gained popularity due to the fact that they don't required labeled data. These methods model the distribution of normal samples and then identify anomalous ones via finding outliers. The main limitation of unsupervised methods is that they make the assumption that anomalies are located in low-density regions and therefore their performance is highly dependant on the alignment of these assumption and the underlying anomaly type.

To detect anomalies and outliers in semi supervised way,  we propose a method that combines a deep auto encoder  and feed forward perceptrons (FFP). This new method that we called \textbf{AnoRand}, jointly optimizes the deep autoencoder and the FFP model in an end-to-end neural network fashion. AnoRand has two building blocks: a Feed Forward Perceptron block and an autoencoder block. The inspiration for this method comes from the fact that when dealing imbalance data, supervised algorithm tend to learn only the majority class. 
The main idea is to learn one class (e.g. the majority class in case of anomaly detection) as well as possible by taking advantage of the ability of auto encoders to represent data in a latent space and the ability of Feed Forward Perceptron (FFP) to learn one class when the data is highly imbalanced.
Indeed, Anand et al. \cite{anand1991improved} showed that in a classification problem, the norm of the gradient vector of a class depends on the size of the class. Thus, in case of imbalance classes, the majority class therefore contributes more to updating the model parameters during gradient descent and at the end the model only learns the majority class.
In our method, the FFP block has a role of informing and reinforcing the capacity of the autoencoder block to embed the normal samples. Our method is performed in two steps: (1) we first create synthetic anomalies by randomly adding noise to few samples from the training data; (2) secondly we train our deep learning model in supervised maner with the new labeled data.
We compared the performance of the proposed method to 17 state-of-the-art unsupervised anomaly detection method on synthetic data sets and 57 real-world data sets from the ADBench benchmark  paper \cite{han2022adbench}. 
Our results show that this new method generally outperforms most of the state-of-the-art methods and has the best performance (AUC ROC and AUC PR) on the vast majority of reference datasets. 
In particular, AnoRand outperforms Deep auto encoder, Variational auto encoder and MLP even though they have the same kind of building blocks.
 We also tested our method (AnoRand) in a supervised way by using the actual labels to train the model instead of creating synthetic label as we did in the semi supervised case. The results show that it has very good performance compared to most of state-of-the-art supervised algorithms. Our results also show that classical methods such as SVM, CatBoost, LGB tend to have better performance than most of deep learning based algorithms.
The main contributions
of our paper are:
\begin{itemize}[leftmargin=*]
\item We propose a new anomaly detection method called AnoRand that does not require any assumption about the shape of the decision boundary between the normal samples and anomalous ones.
\item  We learn a decision boundary using only the normal samples and noisy version of few of them.
\item We compared AnoRand performance to 17 state-of-the-art unuspervised algorithm and 57 real-world anomaly detection benchmark datasets. The results show that it has state-of-the-art performance on most of them.
\item  We tested AnoRand in a supervised way by using the actual labels instead of creating synthetic labels. The results show that it achieves state-of-the-art performance.
\end{itemize}

\section{Related works}
\textbf{Anomaly detection.} 
In data science and machine learning, anomaly detection or outlier detection is generally defined as the identification of rare events or samples which deviate significantly from the majority of the data and do not conform to a well defined notion of normal behaviour.
There are mainly four types of anomalies : (i) Local anomalies when the anomalies are deviant from their local neighborhoods. (ii) Global anomalies are more different from the normal data, generated from a uniform distribution, where the boundaries are defined as the min and max of an input feature. (iii) Dependency anomalies refer to samples that do not follow the dependency structure which normal data follow. (iv) Clustered anomalies, also known as group anomalies, exhibit similar characteristics. In the literature, there are three group of anomaly detection algorithms: Supervised, semi-supervised and unsupervised algorithms.

\subsection{ Anomaly Detection Algorithms}
\textbf{Unsupervised algorithms.}
Unsupervised methods make the assumption that anomalies are located in low-density regions. They are effective when the input data and the algorithm assumption(s) meet. These algorithms can be grouped into 2 categories: classical algorithms and deep learning based algorithms. (i) Classical algorithms include  distribution and distance based unsupervised methods. Most popular algorithms include: K Nearest Neighbors (KNN) \cite{ramaswamy2000efficient};
Isolation Forest (IForest) \cite{4781136};
One-class SVM (OCSVM) \cite{scholkopf1999support} ; The Empirical-Cumulative-distribution-based Outlier Detection (ECOD) \cite{li2022ecod}. There are other unsupervised methods of this category such as: Local Outlier Factor (LOF) \cite{breunig2000lof}, Cluster-based Local Outlier Factor (CBLOF) \cite{he2003discovering}, Connectivity-Based Outlier Factor (COF)\cite{tang2002enhancing},Histogram- based outlier detection (HBOS) \cite{goldstein2012histogram}, Subspace Outlier Detection (SOD) \cite{kriegel2009outlier},  Copula Based Outlier Detector (COPOD) \cite{li2020copod}, Principal Component Analysis (PCA) \cite{shyu2003novel}, Lightweight on-line detector of anomalies (LODA) \cite{pevny2016loda}. More details and python implementation of these algorithms can be found in Han et al \cite{han2022adbench} and the python package Pyod \cite{zhao2019pyod}.
(ii) Deep learning based algorithms includes algorithm that use deep learning representation to cluster data into homogeneous classes.
Deep Support Vector Data Description (DeepSVDD) \cite{ruff2018deep} uses the idea of OCSVM by training a neural network to learn a transformation that minimizes the volume of a hyper-sphere in the output space that encloses the samples of one class. All the samples that are far from the center of the hyper-sphere are labeled as anomalies. Deep Autoencoding Gaussian Mixture Model (DAGMM) \cite{zong2018deep} optimizes jointly a deep autoencoder and a Gaussian mixture model in a same learning loop. Auto encoders have also been used for anomaly detction by some authors \cite{kingma2013auto,zhou2020encoding,zavrtanik2021draem,shi2021unsupervised}. They used the reconstruction error as anomaly score.
It has been shown \cite{collin2021improved,you2022a} that reconstruction based anomaly detection method such as auto encoders lead to high number of false alarms. Indeed, auto encoder tend to produced blurry output. This behavior can lead to a kind of smoothing or blurring the anomalies such that will look like normal samples.

\textbf{Fully supervised and Semi-supervised algorithms.}
These methods are designed to use few label during training. Fully supervised methods require full access to labeled data. They can be very effective in detecting known anomalies but may fail to detect unknown ones. Supervised methods may also perform poorly on imbalance data. Indeed when the classes are imbalanced, the algorithms tend to produce high rate of false negative. The algorithms are biased and fail to learn the minority class. To deal with this issue, there are two main approaches : the resampling approach  which consists in balancing the classes by adding or removing data from classes  and the approach which consists in modifying the learning algorithms so that they take into account the class imbalance. These algorithms include classical methods such as SVM, random forest \cite{zhao2018xgbod,bayes1763lii,cortes1995support,breiman2001random,chen2016xgboost} and deep learning based methods \cite{akcay2019ganomaly,ruff2020deep,pang2018learning,pang2019deep,pang2019deep,zhou2021feature,rosenblatt1958perceptron,gorishniy2021revisiting}.

\section{AnoRand method and implementation details}

\subsection{Proposed Architecture}
Lets first define  $(x,y)=\{ (x_{1},y_{1}),(x_{2},y_{2}),\dots (x_{N},y_{N}) \}$ in the machine learning framework such that $y^{i}_{1}$ is the target (label) vector and $x_{i} \in R^{d}$ the feature vector for the $ith$ sample. 

In this paper we proposed to combine an autoencoder architecture with a fully connected architecture to detect outliers and anomalies. The proposed architecture has two building blocks: (1) a noise detection (ND) block composed of Feed Forward Perceptron (FFP) and (2) an autoencoder (AE) block. We call this new method \textbf{AnoRand}. AnoRand jointly optimizes the deep autoencoder and the FFP model in an end-to-end neural network fashion. 
The joint optimization balances autoencoding reconstruction and helps the autoencoder escape from less attractive local optima and further reduce reconstruction errors. 
The main idea of this new method is to learn one class (e.g. the majority class in case of anomaly detection) as well as possible by taking advantage of the ability of auto encoders to represent data in a latent space and the ability of Feed Forward Perceptron (FFP) to learn one class when the data is highly imbalanced. Indeed, Anand et al. \cite{anand1991improved} showed that in highly imbalance classification problem, the expectation of the total gradient is dominated by that of the majority class. The majority class therefore contributes more to updating the model parameters during gradient descent. Our method takes advantage of these limitations of FFP models.
AnoRand method is performed in two steps: First, we create synthetic anomalies by randomly disturbing (add noise) few samples (e.g. 2\%)  from the training set. Second, we use the normal and the synthetic samples as input to our model. The synthetic label generation is describe in subsection \ref{labelgen}.
In this architecture, the noise detection block has a role of informing and reinforcing the capacity of the autoencoder block to embed the normal samples. We call the FFP block a noise detection block because our experiments show that when the synthetic labels are generated using high value of noise, the FFP block outputs high value for anomalies.

 Lets denote by $z_{1}=F(x,\theta_{0})$ the mapping of the input features by FFP and $z_{0}=E(x,\theta_{1})$ its mapping by the encoder from an auto encoder architecture. Where $\theta_{1}$ is the weights of the Feed Forward Perceptron (FFP) and $\theta_{0}$ the weights of the auto encoder block.
The final latent vector of the autoencoder block is define as 
$    z=(z_{0},z_{1})=(F(x,\theta_{0}),E(x,\theta_{1}))$.
The final model has one input and tree outputs. The model takes as input the features X and the some synthetic labels Y generated as described in subsection \ref{labelgen}. The first output $\hat{y}_{1}$ is the probability of the sample being a noisy sample estimated by the noise detection block, the second output $\hat{x}$ is the reconstructed signal of the autoencoder block and the third output $\hat{y}_{0}$ is the predicted probability of the sample being an anomaly estimated by the AE block. Notice that $\hat{y}_{0}$ is computed using the reconstruction error of the sample. 
We make the hypothesis that as the model learns the normal class, when the reconstruction error is high, the sample is more likely to be an anomaly.
The final model prediction is a weighted sum of $\hat{y}_{1}$ and $\hat{y}_{0}$. The full network architecture is described in figure \ref{fig:my_arch}.

\usetikzlibrary {shapes.geometric}
\newcommand\drawNodes[2]{
  \foreach \neurons [count=\lyrIdx] in #2 {
    \StrCount{\neurons}{,}[\lyrLength] 
    \foreach \n [count=\nIdx] in \neurons
      \node[neuron] (#1-\lyrIdx-\nIdx) at (2*\lyrIdx, \lyrLength/2-1*\nIdx) {\n};
  }
}
\begin{figure}
\begin{adjustbox}{height=8cm,width=\textwidth}
\begin{tikzpicture} [ae/.style={trapezium, draw, minimum height=1cm, minimum width=4cm},
neuron/.style={circle, draw, minimum size=4ex, thick,fill=yellow!10},
squarednode/.style={rectangle, draw=red!60, fill=red!5, very thick,minimum height=6cm, minimum size=0.8cm},
node 1/.style={nodeIn},]
\draw[orange, very thick,dotted] (-5,-1) rectangle (12.5,9.5);
\node (mlp)[ae,fill=blue!40,dotted,ultra thick, shape border rotate=240,scale=0.8] at (0,7.5) {FFP $F(x,\theta_{0})$};
\node (full)[squarednode,fill=blue!40] at (5.5,5.5) {Dense};
   \node (pred)[squarednode] at (8,7.5) {$\hat{y}_{1}$};
\node (AEpred)[squarednode] at (5,0) {$\hat{y}_{0}=\frac{1}{1+e^{-(x-\hat{x})^2}}$};  
\node (pred0)[ squarednode] at (8,0) {$\hat{y}_{0}$};
   \node (Encoder)[ae,fill=red!40, shape border rotate=240,scale=0.8] at (0,4) {Encoder $E(x,\theta_{1})$};
   \node (latentMlp)[squarednode,fill=blue!40] at (3.5,6) {$z_{1}$};
   \node (latent)[squarednode] at (3.5,5) {$z_{0}$};
   \node (decoder)[ae,fill=red!20, shape border rotate=60,scale=0.8] at (8.5,4) {Decoder $D(x,\theta_{1})$};

\def\NI{5} 
\def\NO{5} 
\def\yshift{0.4} 

\foreach \i [evaluate={\c=int(\i==\NI); \y=\NI/2-\i-\c*\yshift; \s=\NI/2-\i-\c*\yshift; \index=(\i<\NI?int(\i):"n");}] in {1,...,\NI}{ 
\node[,outer sep=0.6] (NI-\i) at (-3,\y+6.5) {$x_{\index}^{(0)}$};
\node[,outer sep=0.4] (NO-\i) at (11,\s+4.6) {$\hat{x}_{\index}$};
}
  \path (NI-\NI) --++ (0,1+\yshift) node[midway,scale=1.2] {$\vdots$};
  \path (NO-\NO) --++ (0,1+\yshift) node[midway,scale=1.2] {$\vdots$};

\node (inputs) [label=\textbf{Inputs}, fit=(NI-1) (NI-5), draw, fill=blue!10, opacity=0.3] (input) {};
\node (AEout) [label=\textbf{\tiny Reconstructed}, fit=(NO-1) (NO-5), draw, fill=green!20, opacity=0.3] (AEout) {};

      \draw[->,dashed,red, thick] (input.east) to [out=-10,in=160] (mlp.west);
    \draw[->,dashed,red, thick] (input.east) to [out=-10,in=160] (Encoder.west);
  \draw [->,red,dashed,ultra thick] (decoder) -- (AEout);
  \node [label=\textbf{Concat}, fit=(latentMlp) (latent), draw, fill=blue!10, opacity=0.3] (concat) {};
   \draw [->,red,dashed,ultra thick] (latentMlp.east) to [out=-10,in=160] (pred.west); 
   
    \draw[->,dashed,blue, thick] (mlp.east) to [out=-10,in=160] (latentMlp.west);

    \draw[->,dashed,red, thick] (Encoder.east) to [out=-10,in=160] (latent.west);
    \draw[->,dashed,red, thick] (full.east) to [out=-10,in=160] (decoder.west);
    \draw[->,dashed,blue!20, thick] (input.south) |-(AEpred.west); 
    
    \draw[->,dashed,blue!20, thick]  (AEout.south)  to [out=-30,in=100] (AEpred.north); 
    \draw [->,red,dashed,ultra thick] (concat) -- (full);
    \draw [->,red,dashed,ultra thick] (AEpred) -- (pred0);  
\end{tikzpicture}
\end{adjustbox}
    \caption{Proposed architecture}
    \label{fig:my_arch}
\end{figure}
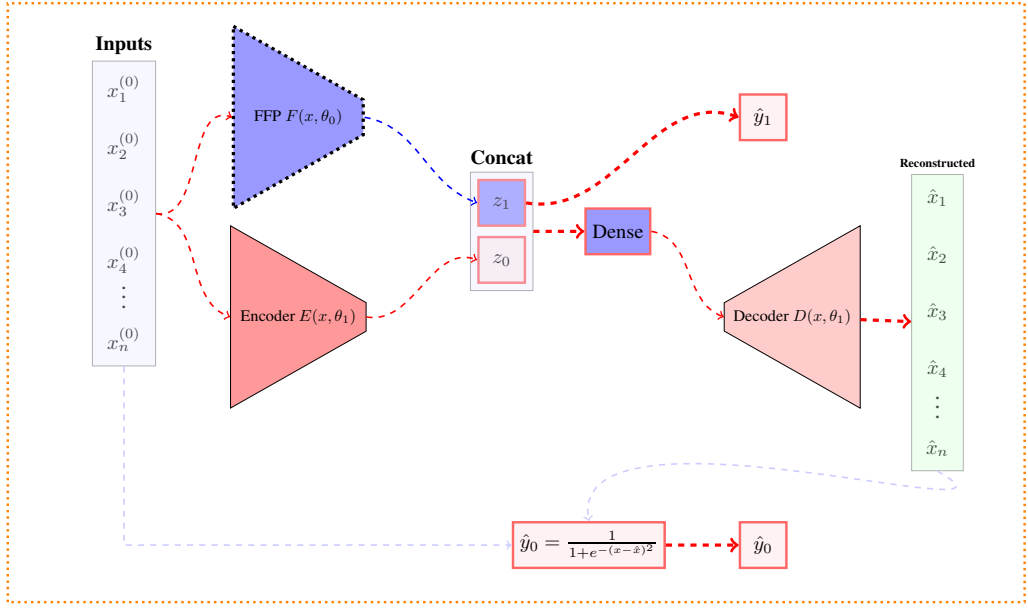

\subsection{Synthetic label generation}
\label{labelgen}

Lets define $X_{0}$ the samples from the normal class labeled 0. To build our semi supervised model, we created the synthetic anomalies (fake anomalous samples) as follows: we first selected randomly few samples (for example 1\% of $X_{0}$) from the normal class, we denoted these samples as $X_{0}^{0}$ and we refer to the remaining samples as $X_{1}$. Secondly, we added some noise to $X_{0}^{0}$ and label them as anomalous. The noisy samples are referred to as $X_{0}^{1}$. The noisy samples can be generated by using Gaussian noise or other sophisticated methods such as in Cai et al.\cite{cai2022perturbation}. 
In our experiment, we combined Gaussian noise with the synthetic minority over-sampling technique (SMOTE) introduced by Chawla et al.\cite{chawla2002smote} to create these noisy samples. This technique consists creating synthetic sample by using some k  nearest neigbord of a sample in the minority class. For a given sample in the minority class, we randomly select k neighbors and generate one sample in the direction of each one. Each synthetic sample is generated by taking the  difference between the feature vector (sample) under consideration and its nearest neighbor and multiply this difference by a random number between 0 and 1. The obtained value is then added to the feature vector of the sample under consideration. In our experiments, we first labeled $X_{0}^{0}$ as anomalous and then oversampled them using SMOTE up to 5\% of the total training dataset. Then added a Gaussian noise to $X_{0}^{0}$. By combining the SMOTE technique with Gaussian noise we make sure that the synthetic anomalies are not very far from the normal samples in terms of distribution. This is important in real world application because anomalies are just modified or "broken" versions of the normal samples. 
So the final training dataset $X_{tr}=X_{1}+X_{0}^{0}+X_{0}^{1}$ is composed of 5\% of "anomalous" samples. 

\subsection{Objective function}
The final loss of the model is composed of two parts: the cross entropy computed using the noise detection block and the cross entropy obtained by the auto encoder reconstruction error.
The loss function is defined as follows:
\begin{equation}
   \mathcal{L}(\theta) =  w\cdot \mathcal{L}(\theta_{1}) +  (1-w)\cdot \mathcal{L}(\theta_{0})
\end{equation}
\begin{equation}
    \mathcal{L}(\theta_{0})=\sum_{i=1}^{N} \mathcal{L}(z(x_{i},\theta_{0}),y_{i}) =-\sum_{i=1}^{N}y_{i}\log(y^{i}_{0})+(1-y_{i})\log(1-y^{i}_{0}) 
\end{equation}

\begin{equation}
 \mathcal{L}(\theta_{1})=\sum_{i=1}^{N} \mathcal{L}(z(x_{i},\theta_{1}),y_{i}) =-\sum_{i=1}^{N}y_{i}\log(y^{i}_{1})+(1-y_{i})\log(1-y^{i}_{1})
\end{equation}

Where $\theta=(\theta_{0},\theta_{1})$ and 
$w \leq 1$.
$y^{i}_{0}=z(x_{i},\theta_{0})$ is the estimated probability of sample $ i $ being a noisy sample estimated by the ND block and $y^{i}_{1}=z(x_{i},\theta_{1})$ is the estimated probability of sample $ i $ being an anomaly computed using the model final prediction. Recall that $\theta_{0}$ and $\theta_{1}$ are respectively some adjustable parameters of the ND block and the AE block.
$\mathcal{L}(\theta_{0})$ is the cross entropy of the ED (autoencoder) block and $\mathcal{L}(\theta_{1})$ is the cross entropy of the ND (noise detector) block.
N is the total number of samples. The final loss function is a weighted sum of the cross entropy losses. 
Note that $\mathcal{L}(\theta_{0})$ is computed using the reconstruction error. We look for the optimal value of $w$ by varying it between 0 and 1. The higher $w$ is, the greater the role of the ND block in the final prediction. In all our experiments, we set empirically $w=0.2$ (see subsection \ref{choiceW1}).
To train our model, we try to find a configuration of the model parameters that minimizes the training loss 
\begin{equation}
    \hat{\theta}= \operatorname*{arg\,min}_\theta \mathcal{L}(\theta)=\operatorname*{arg\,min}_\theta \sum_{i=1}^{N} \mathcal{L}(z(x_{i},\theta),y_{i})
\end{equation}
Note that, $\theta_{0}$ and $\theta_{1}$ are not mutually independent or exclusive. Indeed, $\mathcal{L}(\theta_{0})$ and $\mathcal{L}(\theta_{1})$ are computed separately but during the back-propagation process, some parameters of the model are updated using the gradients of $\mathcal{L}(\theta_{0})$ and $\mathcal{L}(\theta_{1})$ with respective weights $1-w$ and $w$.

\subsection{Evaluation Metrics}
We evaluate the algorithms by using two widely used metrics: AUC ROC (Area Under Receiver Operating Characteristic Curve) and AUC PR (Area Under Precision-Recall Curve). The AUC PR shows precision values for corresponding recall values. It provides a model-wide evaluation like the AUC ROC plot. 
The AUC PR measures the entire two-dimensional area under the entire precision-recall curve (by integral calculations) from (0,0) to (1, 1).
In all incoming experiments, we report these two metrics as performance metrics. The higher the values, the better is the algorithm.
To compare the algorithms, we will use  AUC PR metric instead of the AUC ROC  as Saito et~al. \cite{saito2015precision} show in their study, the AUC ROC may not be well suited in case of highly imbalanced classes.  In their article \cite{saito2015precision}, these authors showed that AUC ROC could be misleading when applied in imbalanced classification scenarios instead AUC PR should be used. 

\subsection{Anomaly score}
To compute the model final prediction, we combine the output of the noise detection block and the output of the autoencoder block. We call the FFP block a noise detection block because our experiments show that when the synthetic labels are generated using high value of noise, the FFP block outputs ($\hat{y}_{1}$) high value for anomalies. This means that both $\hat{y}_{0}$ and $\hat{y}_{1}$ have high values for anomalies or outliers. So to be sure that we catch all anomalies and outliers, we combine these two values to compute the final anomaly score.
The final prediction is then a weighted sum of $\hat{y}_{1}$ and $\hat{y}_{0}$. Lets define $\alpha$ the weight of the the autoencoder block output $\hat{y}_{0}$ inside the weighted sum.
We first compute the third quantile $Q_{3}$ of each output and compute $\alpha$ as follows:

\begin{equation}
    \alpha=\frac{Q_{3}^{1}}{Q_{3}^{0}+Q_{3}^{1}}
\end{equation}
Where $Q_{3}^{0}$ is the third quantile of the the autoencoder block predicted probabilities $\hat{y}_{0}$ and $Q_{3}^{1}$ is the third quantile of the the noise detection block predicted probabilities $\hat{y}_{1}$. We used the third quantile $Q_{3}$ just to make sure that we have a good estimate of the range of the predicted probability. One can use any other quantile or aggregating metrics (mean, variance etc.). The quantiles are more suited because they are less sensitive to extreme values. 
The final model is defined as follows:
\begin{equation}
    \hat{y} =\hat{y}_{1}\cdot (1-\alpha)+ \alpha\cdot \hat{y}_{0}
\end{equation}

\section{Experiment}
For all upcoming experiments, we set the hyper-parameters of our model as follows: the FFP block has 2 hidden layers with respectively 32,16 neurons, the Encoder has two hidden layers with respectively 32,16 neurons and final latent layer has 16 neurons. We chose these values arbitrarily and did not do any further hyper parameter optimization to seek for best parameters. 
We did not spend time on hyperparameter optimization because our goal was to show that the proposed method works well even with arbitrary hyperparameters. For the state-of-the-art algorithms we used their implementation in the python  Outlier Detection (PyOD) package \cite{zhao2019pyod}. We set the hyper-parameters to there default values. Our models are trained for 200 epochs on 1 GPU (NVIDIA GetFore 8GB) with batch size 128. The learning rate is $1 \times 10^{-4}$. 
We compared the performances of our method to those of 18 baseline unsupervised clustering algorithms including: CBLOF \cite{he2003discovering},HBOS \cite{goldstein2012histogram}, KNN\cite{ramaswamy2000efficient}, IForest \cite{4781136}, LOF \cite{breunig2000lof}, OCSVM \cite{scholkopf1999support}, PCA\cite{shyu2003novel}, COF\cite{tang2002enhancing}, SOD \cite{kriegel2009outlier}, COPOD \cite{li2020copod}, ECOD \cite{li2022ecod}, AutoEncoder \cite{kingma2013auto}, DeepSVDD \cite{ruff2018deep},  GMM \cite{zong2018deep} and LODA\cite{pevny2016loda}. These unsupervised algorithms are readily available in the Python Outlier Detection (PyOD) package \cite{zhao2019pyod}. We also added a simple MLP classifier trained using the synthetic labels.

\begin{figure}[!h]
     \centering
     \begin{subfigure}[b]{0.49\textwidth}
         \centering
         \includegraphics[width=\textwidth]{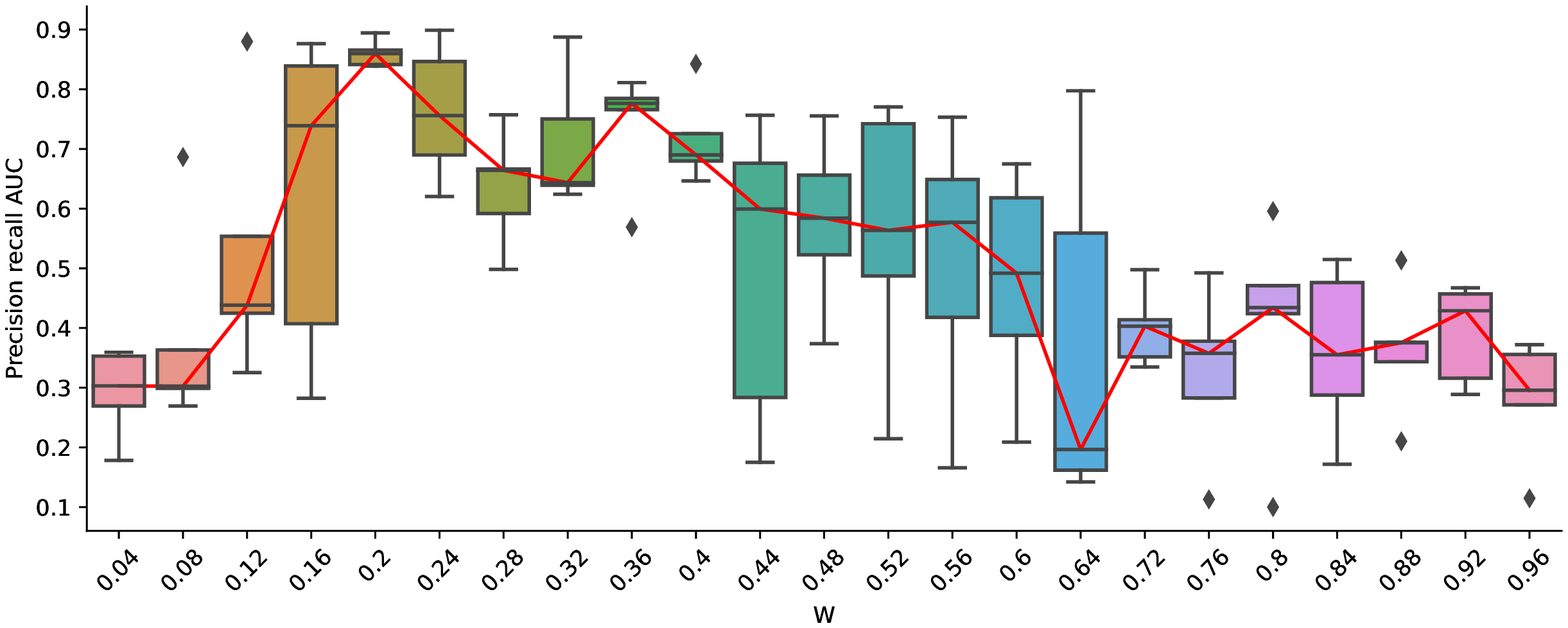}
       \caption{AUC PR}
         \label{fig:w1_pr_auc}
     \end{subfigure}
     \hspace{0.5em}%
     \begin{subfigure}[b]{0.49\textwidth}
         \centering
        \includegraphics[width=\textwidth]{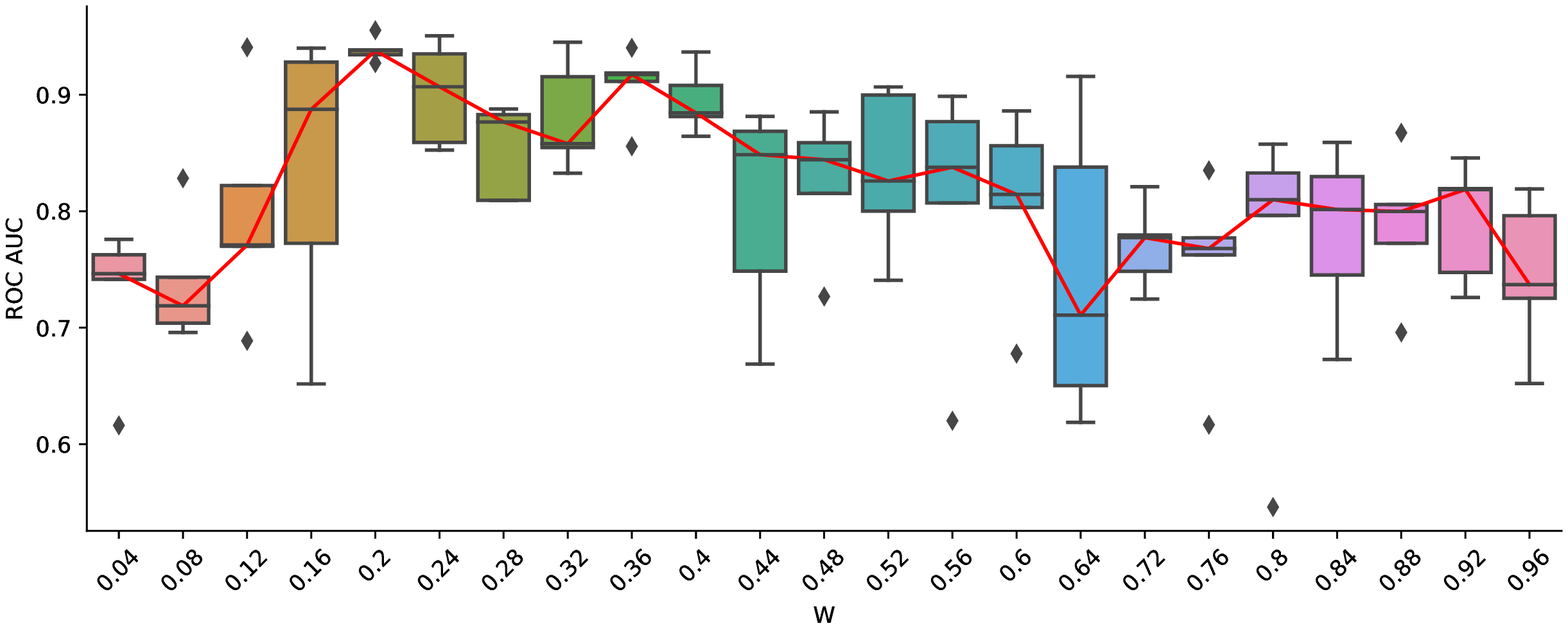}
         \caption{ROC AUC }
         \label{fig:w1_roc_auc}
     \end{subfigure}
     \caption{Performance metrics by varying w}
     \label{fig:w1}

\end{figure}

\textbf{Data generation and splitting.} In these experiments, we simulated a classification dataset using the "make\_classification" module from sklearn. The "make\_classification" module creates clusters of samples normally distributed about vertices of an  hypercube and assigns an equal number of clusters to each class. It then introduces interdependence between the created features and adds various types of noise to the data.
We generated a training set of 20000 samples with an imbalance rate of 5\%. This means that the minority class represents 5\% of the training dataset. Note that for iteration in each experiment, we generated new samples by varying the random state parameter.\\
\textbf{Choice of the optimal value for $w$.}
\label{choiceW1}
Recall that $w$ is the weight assigned to the cross entropy of the noise detection block.
We make the hypothesis that when $w$ tends to 1, the influence of the autoencoder block tends to 0 and the final model is equivalent to a simple Feed Forward Perceptron (FFP) model. So by varying the weights, we expect to see the impact of each part of our architecture to the final model loss.
For each value of $w\in [0,1]$, we trained our model 10 times on 10 different samples and report its AUC PR in figure \ref{fig:w1_pr_auc} and the AUC ROC in figure \ref{fig:w1_roc_auc}. 
The figures show that the model performance increases until 0.2 and decreases very fast when $w$ is greater than 0.2. The boxplot at 0.2 shows, the model's AUC PR and AUC ROC are stable. Indeed, at this point, the interquartile range of the boxplots are small and there are less outliers. These results suggest that in the proposed architecture, the ND block positively contributes to the performance of the final model up to a certain level. The optimal value of $w$ lays around 0.2. 
  \long\def\GRAPH #graphsgraph3 {}%
  \GRAPH graph1
\begin{figure}[!h]
    \centering
    \includegraphics[width=\textwidth]{segmentation.eps}
    \caption{Data segmentation process}
    \label{fig:my_label}
\end{figure}
\ENDGRAPH

\GRAPH graph2
\begin{figure}[!h]
     \centering
     \begin{subfigure}[b]{0.49\textwidth}
         \centering
         \includegraphics[width=\textwidth]{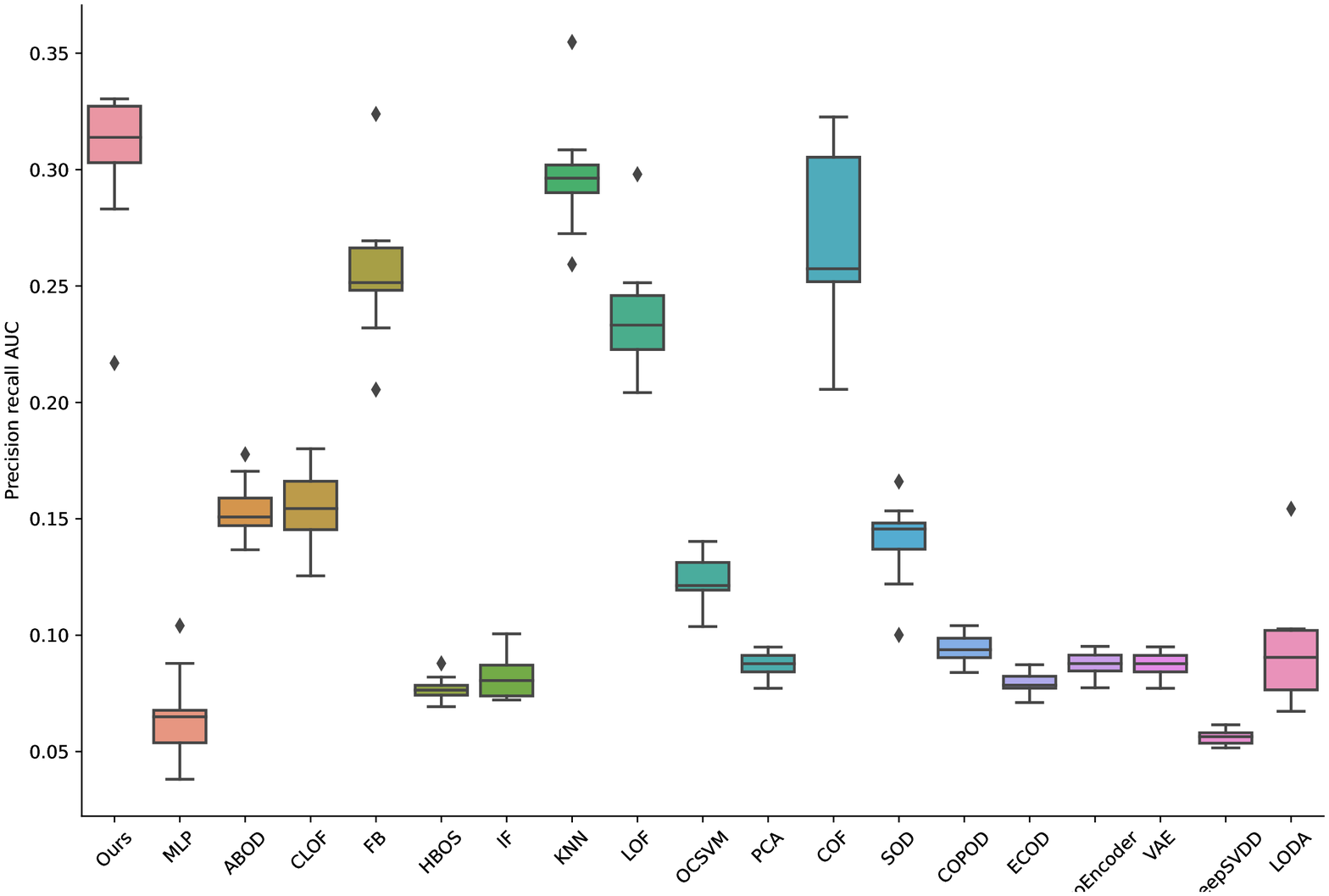}
       \caption{AUC PR }
         \label{fig:synthetic_pr_auc}
     \end{subfigure}
     \hspace{0.5em}%
     \begin{subfigure}[b]{0.49\textwidth}
         \centering
        \includegraphics[width=\textwidth]{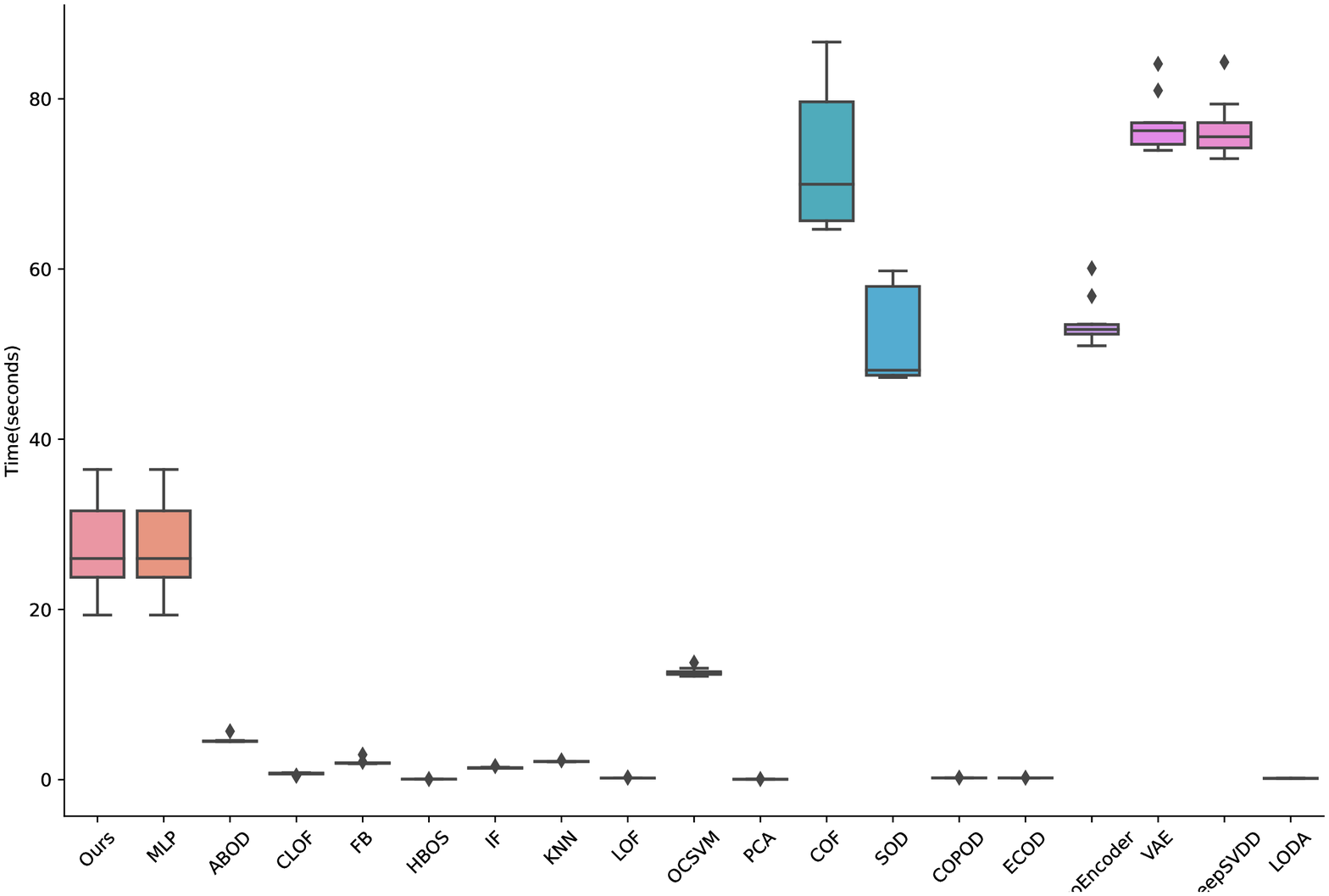}
         \caption{Algorithm Training time in seconds}
         \label{fig:duration}
     \end{subfigure}
     \caption{Performance metrics on synthetic data set for unsupervised algorithms}
     \label{fig:synthetic_perf}

\end{figure}
\ENDGRAPH

\GRAPH graph3
\ENDGRAPH

\GRAPH graph4
\begin{figure}[!h]
     \centering
     \begin{subfigure}[b]{0.45\textwidth}
         \centering
         \includegraphics[width=\textwidth]{images/block_1_output_Pcr_auc_boxPlot_zeros.eps}
       \caption{AUC PR}
         \label{fig:zeros_pr_auc}
     \end{subfigure}
     \hfill
     \begin{subfigure}[b]{0.45\textwidth}
         \centering
        \includegraphics[width=\textwidth]{images/block_1_output_Roc_auc_boxPlot_zeros.eps}
         \caption{AUC ROC}
         \label{fig:zeros_roc_auc}
     \end{subfigure}
     \caption{Performance metrics by disabling some neurons}
     \label{fig:random_zeros}

\end{figure}
\ENDGRAPH


\GRAPH graph5
\begin{figure}[!h]
     \centering
     \begin{subfigure}[b]{0.45\textwidth}
         \centering
         \includegraphics[width=\textwidth]{images/z_varying_imbRate.eps}
       \caption{Precision recall AUC }
         \label{fig:var1}
     \end{subfigure}
     \hfill
     \begin{subfigure}[b]{0.45\textwidth}
         \centering
        \includegraphics[width=\textwidth]{images/z_varying_imbRate_std.eps}
         \caption{Precision recall AUC std }
         \label{fig:var2}
     \end{subfigure}
     \caption{Performance metrics for varying imbalance rate}

\end{figure}
\ENDGRAPH


\GRAPH graph6
\begin{figure}[!h]
     \centering
     \begin{subfigure}[b]{0.49\textwidth}
         \centering
         \includegraphics[width=\textwidth]{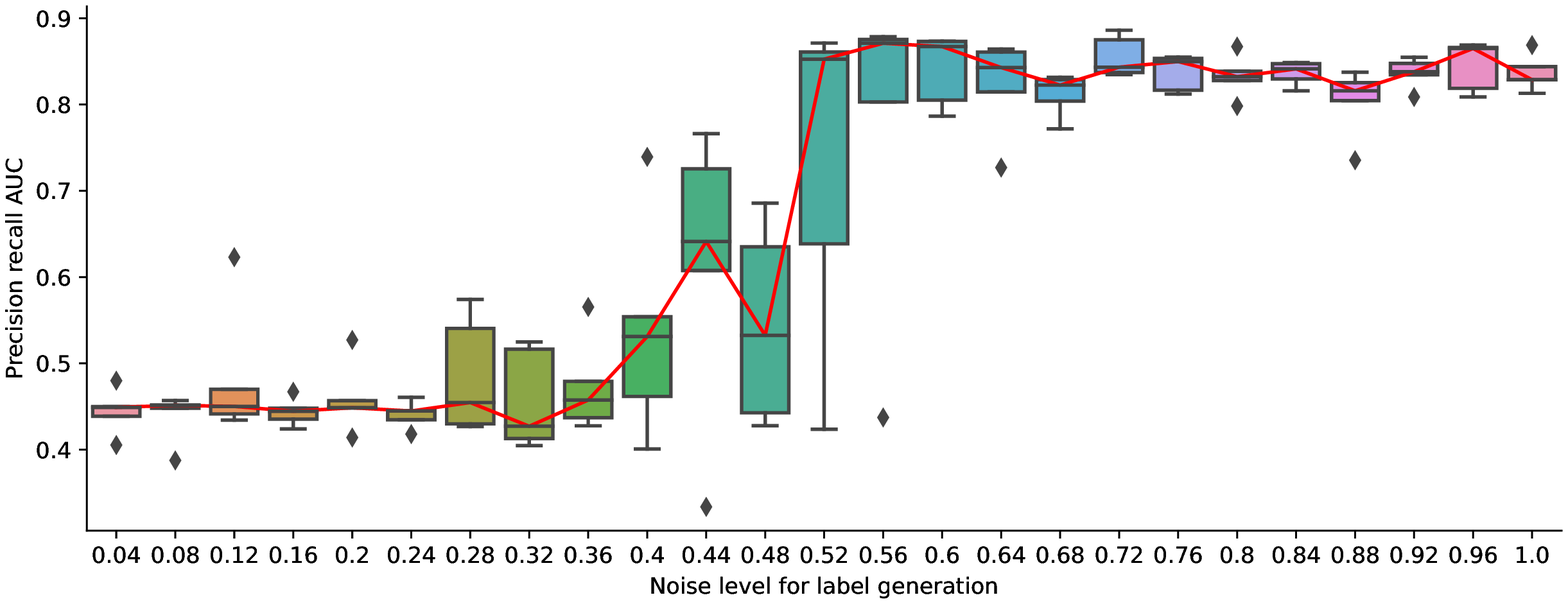}
       \caption{AUC PR}
         \label{fig:pcr_noise}
     \end{subfigure}
     \hspace{0.5em}%
     \begin{subfigure}[b]{0.49\textwidth}
         \centering
        \includegraphics[width=\textwidth]{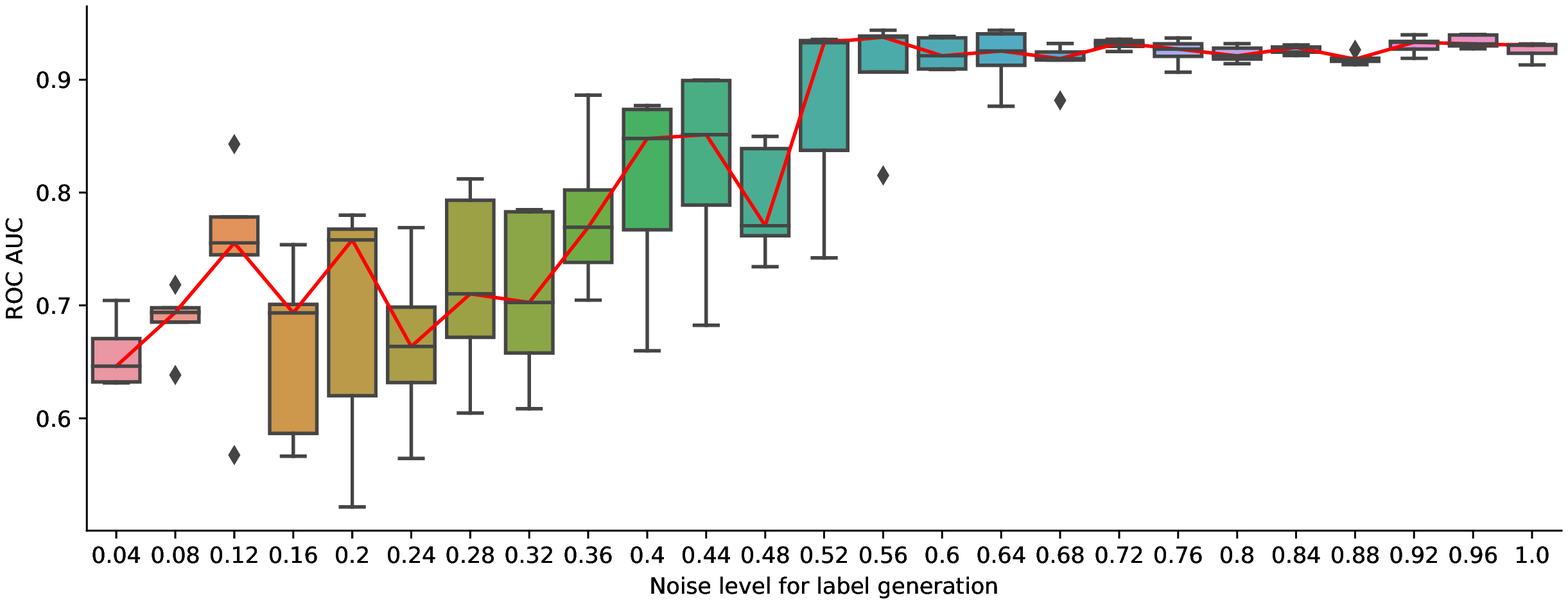}
         \caption{ AUC ROC}
         \label{fig:roc_noise}
     \end{subfigure}
     \caption{Performance metrics for varying noise level when generating labels}

\end{figure}
\ENDGRAPH

\GRAPH graph7
\begin{figure}[!h]
     \centering
     \begin{subfigure}[b]{0.45\textwidth}
         \centering
         \includegraphics[width=\textwidth]{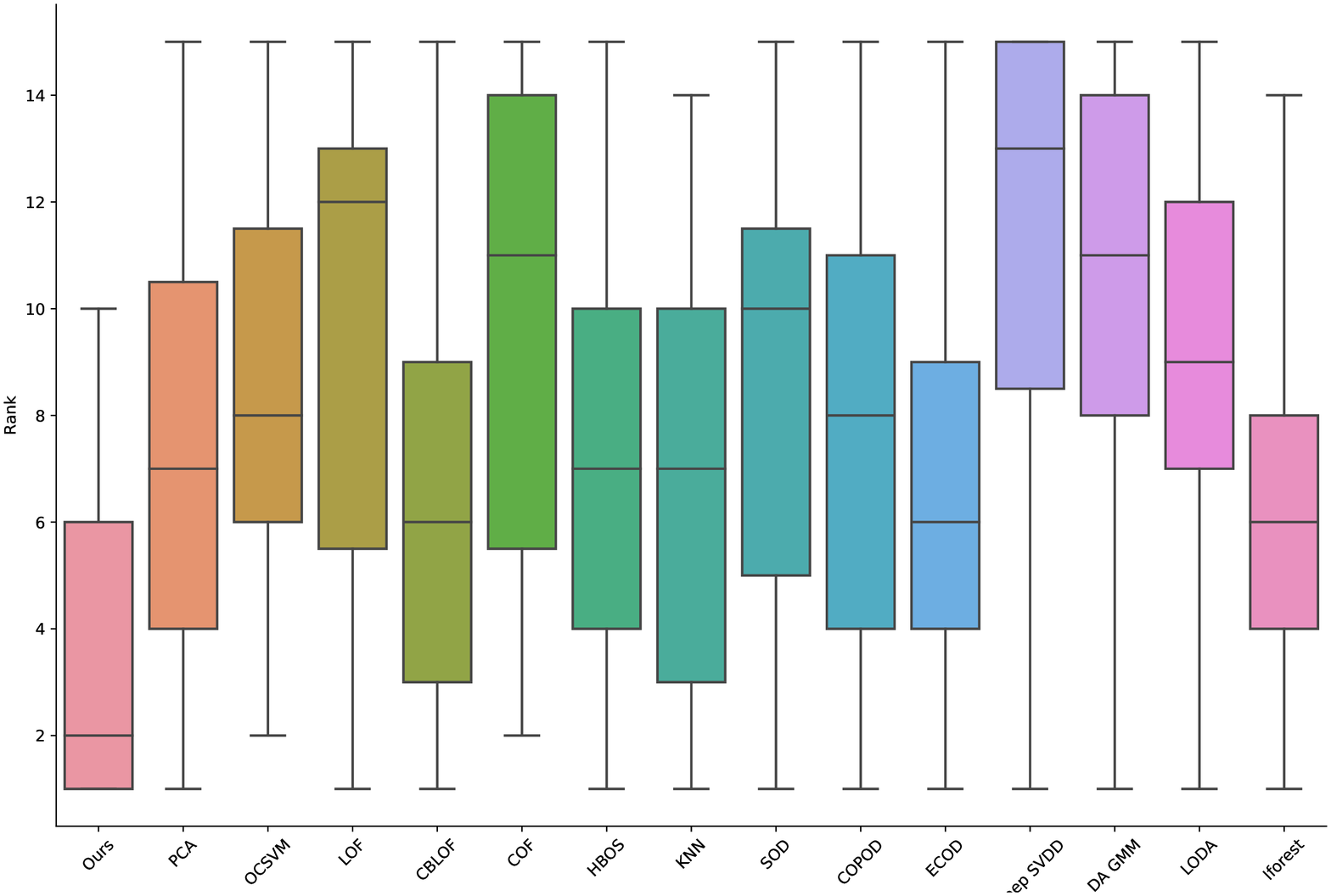}
       \caption{Algorithms rankings (lower the better) }
         \label{fig:adbench}
     \end{subfigure}
     \hfill
     \begin{subfigure}[b]{0.45\textwidth}
         \centering
        \includegraphics[width=\textwidth]{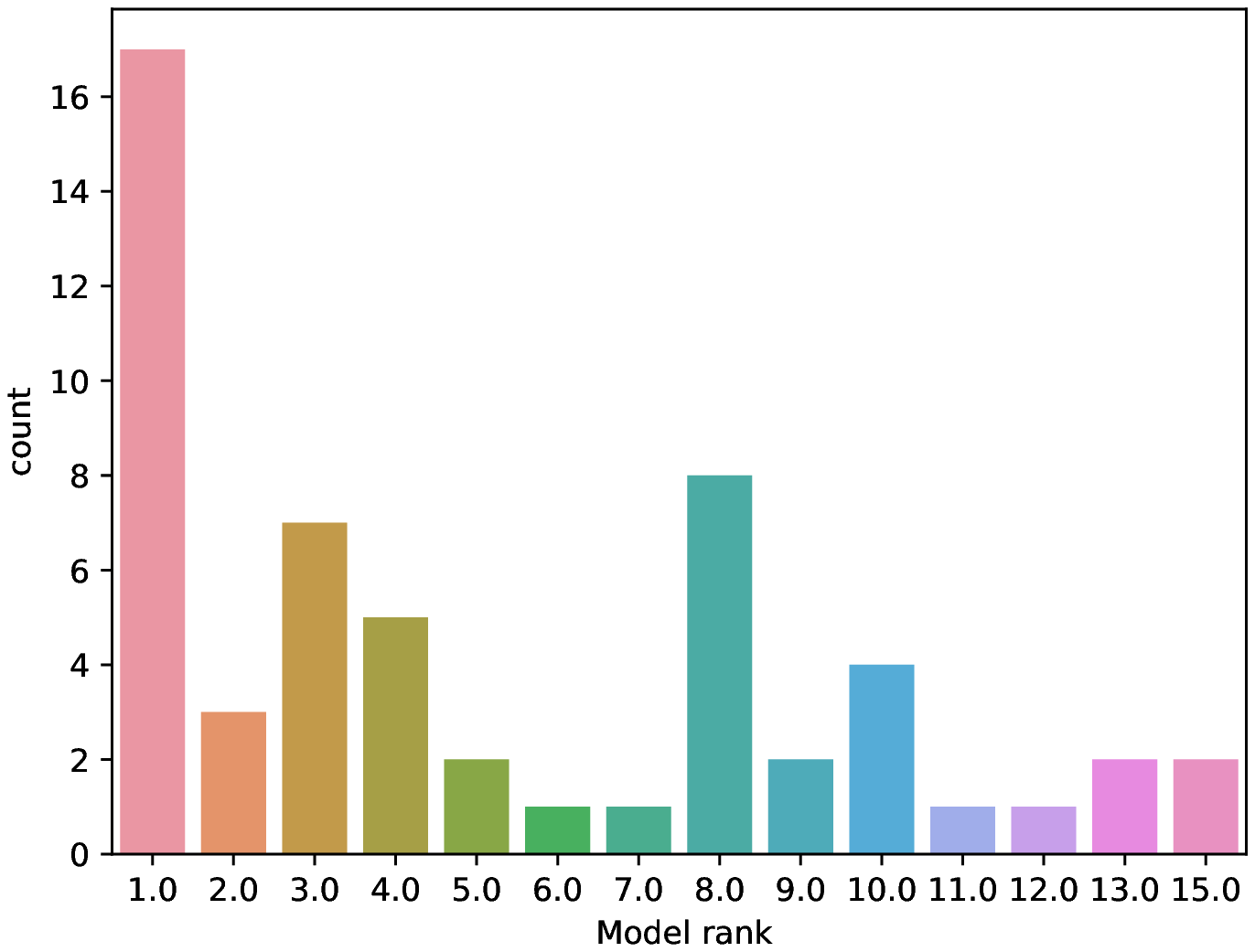}
         \caption{\textbf{AnoRand} (Ours) rankings }
         \label{fig:rank}
     \end{subfigure}
     \caption{Algorithms rankings on real-world data sets}
    \label{fig:Unsupranks}
\end{figure}
\ENDGRAPH

\GRAPH graph8
\begin{figure}[!h]
     \centering
     \begin{subfigure}[b]{0.49\textwidth}
         \centering
         \includegraphics[width=\textwidth]{images/ADBench_ranks.eps}
       \caption{Unsupervised rankings (lower the better) }
         \label{fig:adbenchUnsup}
     \end{subfigure}
     \hspace{0.4em}%
     \begin{subfigure}[b]{0.49\textwidth}
         \centering
        \includegraphics[width=\textwidth]{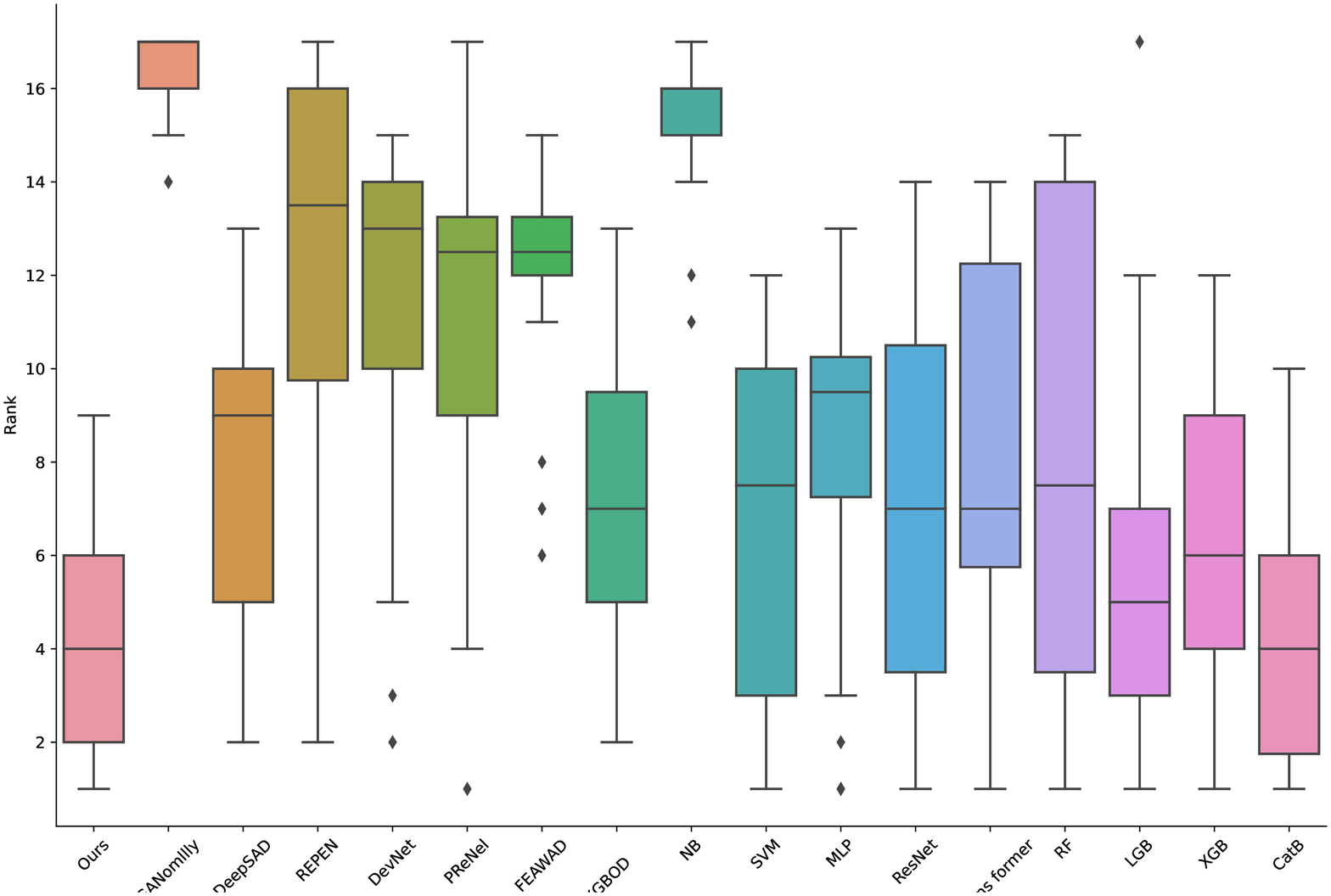}
         \caption{Supervised rankings (lower the better) }
         \label{fig:adbenchSup}
     \end{subfigure}
     \caption{Algorithms rankings on real-world data sets}
    \label{fig:ranks}
\end{figure}
\ENDGRAPH

\GRAPH graph9
\begin{figure}[h]
\begin{center}
\centerline{\includegraphics[width=1\textwidth]{images/ADbenc_SupSIze1.png}}
	\caption{\% of performance change ($\Delta$ ROC PR) between our model's performance and that of the best ste-of-the-art algorithm. Positive value means that our model has the performance and negative value means that another algorithm is better.}
	\label{variation}
\end{center}
\end{figure}
\ENDGRAPH
 
\GRAPH graph10
\begin{figure}[!h]
\begin{center}
\centerline{\includegraphics[width=0.7\textwidth]{images/duration.eps}}
	\caption{Algorithm Training time.}
	\label{duration}
\end{center}
\end{figure}
\ENDGRAPH

\textbf{Noise level when generating synthetic labels.}
Recall that the first step of the proposed method is to generate synthetic labels by introducing some noise inside a very small subset of the normal samples as explained in subsection \ref{labelgen}. These noisy sample will then be considered as the abnormal sample during training. In this subsection we evaluate the impact of the noise level on the models final performance. The final goal is to find out if the amount of noise has an impact on the model final prediction. Lets denote the noise level as the standard deviation of the Gaussian distribution used to create the noisy samples.
Figure \ref{fig:pcr_noise} and \ref{fig:roc_noise} show the model final performance according to the noise level used to generate the synthetic labels. In these experiments we trained 10 models for each noise level. These figures show that the model's performance increases with the noise level. When the noise level is less than 0.32, the AUC PR is stable and lies around 45\%. Between 0.32 and 0.52, the performance increases rapidly but very unstable. When the noise level is greater than 0.52, the model performance becomes more stable and the box plots are more and more small in range. These experiments suggest that the value of the standard deviation of the Gaussian noise should be greater than 0.52 to have better performances. 
  \long\def\GRAPH #graphsgraph6 {}%

\subsection{Anomaly detection on synthetic datasets}
In this subsection, we compared the performance of our architecture to those of some state-of-the-art algorithms on synthetic datasets generated using the "make\_classification" module from python sklearn package. 
Recall that at this stage we select a  small subset (2\%) of the training data to create the synthetic labels (see subsection \ref{labelgen}). 
 We trained and tested each algorithm 10 times and reported the AUC PR and training time in figure \ref{fig:synthetic_perf}. At each iteration, we select randomly 2\% of the samples from the training set and use them to create the synthetic label for our model. By doing so we make sure that the model performance does not depend on the samples used during label generation.

We can see from these two figures that our model outperforms all other models in terms of PRC AUC. In particular, our model outperforms Deep auto encoder, Variational auto encoder and MLP even though they have the same kind of building blocks. We also note that deep learning based unsupervised methods like DeepSVDD and Autoencoder are surprisingly worse than classical methods.
Figure \ref{fig:duration} shows that our method takes more time to train compared to the other algorithms.
  \long\def\GRAPH #graphsgraph2 {}%

\subsection{Unsupervised anomaly detection on real world datasets}
We compared the performance of our method (\textbf{AnoRand}) to those of some state-of-the-art unsupervised methods on the ADBench anomaly detection benchmark
 \cite{han2022adbench}. In there paper, Han et al. \cite{han2022adbench} compared the performances of 14 algorithms on 57 benchmark datasets. The datasets cover different fields including healthcare, security, and more. 
 We grouped the datasets into four categories to make the comparison easy: NLP datasets, Healthcare datasets, Science datasets and datasets from other fields (documents, web etc.).
 In there paper, the authors compared supervised, semi-supervised and unsupervised anomaly detection methods on these datasets. In our study, we only focus on the unsupervised algorithms of the benchmark. 
 In table \ref{real_world_all}  and figure \ref{fig:ranks}, we report the algorithms performance and their rankings on the the ADBench real-world datasets. In figure \ref{fig:adbenchUnsup}, the boxplots show that our model has best ranking among all its counterpart unsupervised algorithms.  These results show that our model has best overall ranking among the tested algorithms. Indeed, figure \ref{fig:adbenchUnsup} hows that AnoRand is ranked first ($1^{st}$) on 22 datasets, second ($2^{nd}$) on 5, third ($3^{rd}$) on 6 and fourth ($4^{th}$) on 4. The results also show in situations where another algorithm outperforms ours, the performance gap is very small in most cases. \\
\textbf{Results on image classification datasets.}
This category includes datasets from computer vision (CV) and image classification datasets such as Mnist, CIFAR10 and MVTec. Pre-trained ResNet18 \cite{he2016deep} models have been used to extract data embedding from the original images.
Table \ref{real_world_all}  shows the algorithms performance on the images benchmark datasets. We can see that \textbf{AnoRand} outperforms all reference algorithms on 5 of the 10 benchmark datasets and has second best performance on 2 other datasets.

\textbf{Results on NLP datasets.}
For NLP datasets, they used a BERT \cite{Devlin2019BERTPO} pre-trained on the BookCorpus and English Wikipedia to extract the embedding of the token. Table \ref{real_world_all}  shows the algorithms performance on 5 NLP benchmark datasets. On these datasets, \textbf{AnoRand} outperforms the other algorithms on the speech and Imdb dataset It is ranked third on the Amazon, Agnews and the Yelp datasets.

\textbf{Results on Healthcare datasets.}
Table \ref{real_world_all}  shows the algorithms performance on the 10 healthcare benchmark datasets. This table shows that \textbf{AnoRand} outperforms the reference algorithms on 3 datasets. It has also the best overall ranking.\\
\textbf{Results on Other datasets.} In this category, we include all other datasets from other fields.
On these datasets, \textbf{AnoRand} outperforms the other algorithm on 10 datasets, ranked second in two and is ranked  third on 2 others. AnoRand has also the best overall ranking.

  \long\def\GRAPH #exp_resultsADBenchAll {}%
\GRAPH ADBenchSup
\begin{table}[!h]
\caption{AUC PR (in \%) of 17 Supervised algorithms. For our method we computed the average AUC PR over 10 runs. The performance rank in parenthesis and best performing method(s) in \textbf{bold}.}
    \centering
    \resizebox{\columnwidth}{!}{%
    \begin{tabular}{lllllllllllllllll|l}
     \toprule
        Datasets & GANomllly & DeepSAD & REPEN & DevNet & PReNel & FEAWAD & XGBOD      & NB & SVM & MLP & ResNet & FTTrans former & RF & LGB & XGB & CatB & \textbf{Ours}  \\ \hline
        campaign & 20.82(16) & 46.45(13) & 15.39(17) & 49.78(12) & 50.98(10) & 37.76(15) & 61.45(2) & 38.67(14) & 51.95(8) & 51.6(9) & 50.33(11) & 54.05(7) & 60.31(4) & 59.83(5) & 58.63(6) & 60.96(3) & \textbf{62.92 $\pm 1.11$(1)} \\  
        celeba & 6.93(16) & 37.04(3) & 3.31(17) & 35.5(5) & 34.49(7) & 26.01(13) & 37.1(2) & 11.6(15) & 32.2(10) & 31.1(12) & 24.68(14) & 33.37(8) & 32.08(11) & 34.64(6) & 33.02(9) & 36.36(4) & \textbf{43.43 $\pm 1.09$(1)} \\  
        cover & 0.92(17) & 98.18(7) & 96.54(14) & 97.44(10) & 97.25(13) & 94.1(15) & 97.63(9) & 92.12(16) & 98.27(6) & 97.36(12) & 98.7(3) & 98.95(2)& 98.1(8) & 98.37(5) & 97.41(11) & 98.45(4) & \textbf{99.55 $\pm 0.08$(1)} \\
        fraud & 42.84(14) & 51.77(9) & 38.4(15) & 59.09(3) & 58.72(4) & 57.48(6) & 45.83(13) & 21.73(16) & 51.19(10) & 54.51(8) & 48.8(11) & 54.99(7) & 62.77(2) & 15.52(17) & 47.34(12) & 58.67(5) & \textbf{73.71$\pm 4.36$(1)} \\ 
        
        CIFAR10 & 9.58(17) & 38.61(10) & 44.06(2) & 39.52(8) & 39.39(9) & 31.04(13) & 40.6(7) & 10.77(16) & 42.5(3) & 38(12) & 38.22(11) & 28.13(14) & 23.98(15) & 42.03(5) & 41.3(6) & 42.28(4) & \textbf{47.67$\pm 11.42$(1)} \\  
        census & 8.46(17) & 48.55(10) & 10.35(16) & 48.11(11) & 49.12(9) & 33.77(14) & 60.7(5) & 10.97(15) & 53.14(7) & 49.39(8) & 42.95(13) & 45.35(12) & 57.07(6) & 61.81(3) & 60.95(4) & \textbf{63.13(1)} & 62.04 $\pm 0.35$(2) \\  
        Waveform & 4.55(17) & 52.55(9) & 22.34(14) & 21.5(15) & 24.61(13) & 32.51(12) & 54.97(7) & 20.4(16) & \textbf{69.28(1)} & 61.47(3) & 50.48(11) & 56.22(6) & 51(10) & 56.86(5) & 53.89(8) & 58.13(4) & 67.20 $\pm 0.17$(2) \\  
        lmdb & 5.05(17) & 35.68(5) & 29.74(10) & 26.93(13) & 27.52(12) & 31.85(8) & 28.69(11) & 9.34(16) & \textbf{48.29(1)} & 46.38(3) & 40.71(4) & 23.94(14) & 11.26(15) & 32.3(7) & 30.83(9) & 32.89(6) & 47.35 $\pm 2.28$(2) \\
        
        magic.gamma & 52.2(17) & 88.07(10) & 77.59(13) & 74.86(15) & 75.47(14) & 80.79(12) & 88.68(8) & 68.16(16) & 87.81(11) & 88.1(9) & 88.77(7) & 89.02(6) & 89.37(5) & 90.06(2) & 89.68(4) & \textbf{90.68(1)} & 89.98 $\pm 0.47$(3) \\
        
        wilt & 4.93(17) & 85.13(10) & 6.7(16) & 8.18(15) & 8.36(14) & 37.94(12) & 93.01(5) & 40.92(11) & 88.15(9) & 34.94(13) & 94.38(2) & \textbf{94.53(1)} & 90.59(8) & 91.83(6) & 90.86(7) & 93.39(4) & 93.66 $\pm 2.09$(3) \\  
        Amazon & 6.08(16) & 35.31(5) & 34.1(6) & 32.76(7) & 32.65(9) & 31.13(12) & 30.1(13) & 5.21(17) & \textbf{48.05(1)} & 45.78(2) & 38.71(4) & 23.17(14) & 9.82(15) & 32.54(10) & 32(11) & 32.71(8) & 42.54 $\pm 1.90$(3) \\  
        FashionMNlST & 23.88(17) & 86.78(3) & 86.09(7) & 84.07(10) & 83.08(11) & 76.64(15) & 85.42(8) & 29.05(16) & 81.81(12) & 85.08(9) & 86.29(5) & 81.79(13) & 76.87(14) & \textbf{87.11(1)} & 86.9(2) & 86.19(6) & 86.38 $\pm 9.72$(4) \\  
        Agnews & 6.34(17) & 75.56(3) & 65.97(7) & 56.15(14) & 57.3(13) & 62.46(11) & 61.84(12) & 8.82(16) & 72.66(5) & \textbf{78.37(1)} & 76.51(2) & 63.03(10) & 29.8(15) & 64.65(9) & 64.83(8) & 66.27(6) & 74.16 $\pm 10.68$(4) \\  
        annthyroid & 45.77(14) & 78.04(11) & 45.07(16) & 45.35(15) & 44.64(17) & 53.95(13) & 93.23(2) & 60.64(12) & 80.97(9) & 79.61(10) & 85.09(8) & 86.5(7) & \textbf{93.25(1)} & 92.97(3) & 92.44(4) & 92.36(5) & 87.04 $\pm 1.68$(6) \\  
        cardio & 53.07(16) & 95.14(11) & 96.27(9) & 92.91(14) & 93.03(13) & 94.6(12) & 98.46(2) & 81(15) & 97.85(3) & 96(10) & \#N/A & \textbf{98.65(1)} & 97.79(4) & 97.4(5) & 96.96(8) & 97(7) & 97.14 $\pm 0.93$(6) \\  
        SVHN & 8.06(16) & 31.84(12) & 35.7(4) & 36.07(2) & \textbf{37.1(1)}& 24.98(13) & 32.39(11) & 5.6(17) & 35.97(3) & 34.03(10) & 35.12(5) & 24.87(14) & 17.52(15) & 34.16(8) & 34.14(9) & 34.45(7) & 34.45 $\pm 12.39$(6) \\  
        
        letter & 16.59(17) & 43.58(12) & 54.56(11) & 32.8(14) & 35.94(13) & 30.17(15) & 71.78(4) & 19.5(16) & 61.32(9) & 56.3(10) & 74.77(2) & 70.47(6) & 69.76(7) & 73.44(3) & 71.06(5) & \textbf{79.14(1)} & 64.04 $\pm 1.74$(8) \\  
        fault & 55.63(17) & 73.77(11) & 69.59(12) & 64.79(14) & 68.92(13) & 64.77(15) & 77.52(6) & 57.44(16) & 74.86(9) & 75.47(8) & 76.6(7) & 79.65(5) & 82.69(3) & 83.41(2) & 81.6(4) & \textbf{83.7(1)} & 74.5 $\pm 1.95$(10) \\  
        MVTee-AD & 57.05(16) & 96.41(7) & 56.48(17) & 88.8(12) & 87.18(14) & 90.87(11) & 97.92(5) & 66.75(15) & 88.38(13) & 95.26(8) & 97.89(6) & 95.13(9) & 98.55(3) & \textbf{98.67(1)} & 98.33(4) & 98.62(2) & 94.11 $\pm 2.38$(10) \\  
        \\ \bottomrule
    \end{tabular}}
    \label{real_world_sup}
\end{table}
\ENDGRAPH


\GRAPH ADBenchAll0
\begin{table}[!ht]
\caption{AUC PR of 15 unsupervised algorithms on 42 real-world datasets. The performance rank is shown in parenthesis (the lower, the better), and mark the best performing method(s) in
\textbf{bold}.}
\label{real_world_all}
    \centering
    \resizebox{\columnwidth}{!}{%
    \begin{tabular}{l|l|llllllllllllll|l}
    \toprule
        Category&Dataset & PCA & OCSVM & LOF & CBLOF & COF & HBOS & KNN & SOD & COPOD & ECOD & Deep SVDD & DA GMM & LODA & Iforest & \textbf{Ours} \\ \midrule
        \multirow{11}{*}{\rotatebox[origin=c]{90}{Image and CV}} & mnist & 39,93(2) & 33,2(4) & 20,9(12) & 28,82(6) & 25,51(9) & 12,51(15) & 35,53(3) & 19,15(14) & 21,35(11) & 31,93(5) & 19,72(13) & 23,75(10) & 25,86(8) & 27,71(7) & \textbf{40,61(1)} \\ 
        & optdigits & 2,76(14) & 2,92(13) & 6,06(4) & 10,08(2) & 4,42(7) & 10,03(3) & 3,06(12) & 4,39(8) & 4,36(9) & 3,43(11) & 2,5(15) & 5,59(5) & 3,95(10) & 5,09(6) &  \textbf{56,34(1)} \\ 
       &  skin & 17,4(12) & 19,03(7) & 18,25(10) & 29,82(2) & 16,38(13) & 23,7(6) & 28,72(3) & 24,61(5) & 17,99(11) & 15,96(14) & 18,48(8) &  NA/NA & 18,44(9) & 26,08(4) &  \textbf{52,2(1)}\\ 
       &  FashionMNIST & 31,42(7) & 31,97(6) & 16,85(14) & 38,9(2) & 20,73(12) & 29,43(9) & 33,87(3) & 28,72(10) & 30,32(8) & 32,53(4) & 17,43(13) & 14,44(15) & 27,32(11) & 32,35(5) &  \textbf{43,16(1)} \\ 
       &  MNIST-C & 16,88(8) & 17,72(7) & 13,84(13) & 27,62(2) & 14,53(12) & 15,46(11) & 22,98(3) & 15,68(10) & 15,9(9) & 18,24(5) & 8,34(15) & 11,37(14) & 18,63(4) & 17,99(6) &  \textbf{35,215(1)} \\ 
       &  satimage-2 & 85,69(4) & 82,71(5) & 4,29(14) &  \textbf{97,09(1)} & 8,8(13) & 78,04(7) & 39,14(10) & 26,11(11) & 76,55(8) & 63,25(9) & 3,08(15) & 22,07(12) & 80,52(6) & 93,45(3) & 94,12(2) \\ 
       &  MVTec-AD & 54,06(9) & 51,44(11) & 54,9(7) &  \textbf{58,52(1)} & 46,59(13) & 55,22(6) & 55,55(4) & 51,48(10) & 54,64(8) & 55,44(5) & 36,5(15) & 45,66(14) & 49,73(12) & 56,04(3) & 57,652(2) \\ 
       &  letter & 6,86(13) & 6,1(15) &  \textbf{34,02(1)} & 14,8(6) & 21,43(5) & 8,38(10) & 30(2) & 28,63(3) & 6,77(14) & 6,94(11) & 9,29(8) & 11,68(7) & 6,87(12) & 8,49(9) & 28,47(4) \\ 
       &  celeba &  \textbf{15,89(1)} & 10,73(6) & 1,71(15) & 11,33(5) & 1,77(14) & 13,82(2) & 3,14(10) & 2,66(11) & 13,69(3) & 12,37(4) & 2,34(12) & 1,95(13) & 4,04(9) & 8,96(7) & 5,61(8) \\  
       &  CIFAR10 & 10,59(6) & 10,19(7) &  \textbf{13,02(1)} & 10,61(5) & 11,61(2) & 8,38(13) & 11,13(3) & 11,06(4) & 8,77(12) & 9,29(10) & 8,05(14) & 7,73(15) & 9,72(9) & 8,97(11) & 10,04(8) \\ \midrule
\multirow{5}{*}{\rotatebox[origin=c]{90}{NLP}}     &  speech & 1,97(11) & 1,96(12) & 2,52(3) & 1,99(10) & 2,25(5) & 2,09(7) & 2,02(9) & 2,13(6) & 1,94(13) & 1,77(15) & 5,12(2) & 2,03(8) & 1,79(14) & 2,31(4) &  \textbf{7,65(1)} \\  
      &   Imdb & 4,55(13) & 4,44(15) & 4,83(6) & 4,75(7) & 5,16(2) & 4,74(8) & 4,49(14) & 4,7(10) & 4,9(4) & 4,9(4) & 5,06(3) & 4,65(11) & 4,59(12) & 4,74(8) &  \textbf{7,65(1)} \\  

& Agnews & 5,74(9) & 5,69(10) &  \textbf{14,35(1)} & 7,02(6) & 12,21(2) & 5,58(11) & 8,61(4) & 8,4(5) & 5,43(12) & 5,43(12) & 4,45(15) & 5,41(14) & 5,93(8) & 6,04(7) & 9,04(3) \\  
         & Amazon & 5,85(10) & 5,64(14) & 5,72(12) & 6,07(5) & 5,74(11) & 5,98(7) & 6,23(2) &  \textbf{6,4(1)} & 6,08(4) & 6,06(6) & 3,84(15) & 5,65(13) & 5,92(9) & 5,95(8) & 6,2(3) \\  
      &   Yelp & 7,62(13) & 7,75(10) & 8,52(5) & 7,68(11) & 8,68(4) & 7,81(9) &  \textbf{9,85(1)} & 9,2(2) & 8,01(6) & 7,98(7) & 6,39(15) & 6,72(14) & 7,65(12) & 7,88(8) & 8,8(3) \\ \midrule
       
\multirow{10}{*}{\rotatebox[origin=c]{90}{Healthcare}} &  WBC & 82,29(7) & 89,87(4) & 5,57(14) & 92,27(2) & 9,73(12) & 73,56(9) & 66,55(10) & 54(11) & 86,19(5) & 86,19(5) & 6,38(13) &  NA/NA & 78,67(8) & 90,49(3) &  \textbf{94,55(1)} \\ 
        &   Cardiotocography & 47,95(4) & 52,61(2) & 30,66(12) & 45,44(5) & 28,21(14) & 38,28(9) & 34,79(10) & 27,99(15) & 40,46(8) & 43,57(6) & 34,03(11) & 30,61(13) & 48(3) & 41,47(7) & \textbf{61,39(1)}\\ 
        &   Lymphography & 97,02(4) & 93,59(5) & 23,08(12) & 97,62(2) & 36,68(11) & 91,83(6) & 38,69(10) & 22,65(13) & 88,68(8) & 90,87(7) & 4,58(15) & 19,52(14) & 44,54(9) & 97,31(3) &  \textbf{99,68(1)} \\ 
        &   breastw & 95,11(7) & 82,7(11) & 28,55(13) & 91,54(9) & 27,6(14) & 97,71(4) & 92,19(8) & 84,88(10) &  \textbf{99,4(1) }& 98,54(2) & 50,92(12) &  NA/NA & 97,04(5) & 96,04(6) & 98,17(3) \\ 
        &   WPBC & 23,01(6) & 22,93(7) & 20,29(15) & 21,32(13) & 21,3(14) & 23,04(5) & 21,49(11) & 25,37(3) & 22,81(8) & 21,38(12) &  \textbf{26,24(1)} & 22,49(9) & 25,58(2) & 22,42(10) & 23,06 $\pm 0.44$(4) \\ 
        &   Hepatitis & 36,65(4) & 29,44(8) & 13,69(15) & 31,54(6) & 14,39(14) & 37,73(3) & 21,95(13) & 24,89(10) &  \textbf{41,5(1)} & 37,82(2) & 22,17(12) & 22,96(11) & 30,9(7) & 26,25(9) & 35,2(5) \\  
        &   thyroid & 44,34(4) & 21,23(10) & 20,81(11) & 29,95(7) & 28,5(8) & 50,98(3) & 34,98(6) & 23,56(9) & 19,64(12) & 54,05(2) & 2,5(15) & 16,06(13) & 14,68(14) &  \textbf{63,11(1)} & 40,6(5) \\ 
        &   annthyroid & 16,12(9) & 10,37(13) & 15,71(10) & 13,69(12) & 14,39(11) & 16,99(5) & 16,74(7) & 18,84(4) & 16,58(8) & 24,65(2) & 21,95(3) & 9,64(14) & 7,06(15) &  \textbf{30,47(1)} & 16,9(6) \\  
        &   Pima & 54,03(5) & 50(8) & 47,18(10) & 53,19(6) & 44,7(11) &  \textbf{56,61(1)} & 55,14(4) & 48,24(9) & 55,19(3) & 37,3(14) & 35,87(15) & 41,55(13) & 44,09(12) & 55,82(2) & 50,09(7) \\ 
        &   cardio & 66,06(2) & 62,89(3) & 23,79(14) & 61,95(4) & 28,67(12) & 52,1(9) & 40,72(10) & 28,54(13) & 60,42(5) &  \textbf{68,59(1)} & 22,5(15) & 28,92(11) & 53,41(7) & 59,95(6) & 52,53(8) \\  \midrule
        
\multirow{18}{*}{\rotatebox[origin=c]{90}{Others}} &  musk & 99,89(4) & 10,61(10) & 2,82(14) & 100(1) & 2,61(15) & 100(1) & 9,65(11) & 7,59(12) & 34,79(8) & 34,95(7) & 5,39(13) & 32,75(9) & 47,6(6) & 99,61(5) &  \textbf{100(1)} \\
      &    Waveform & 5,79(11) & 4,37(14) & 11,33(5) & 18,98(2) & 14,11(3) & 5,86(10) & 13,04(4) & 9,66(6) & 6,9(7) & 6,86(8) & 4,83(12) & 3,11(15) & 4,71(13) & 6,24(9) &  \textbf{33,26(1)} \\ 
 & cover & 9,8(7) & 11,41(5) & 8,12(9) & 5,83(13) & 4(14) & 6,83(11) & 6,16(12) & 3,88(15) & 11,37(6) & 15,63(3) & 8,12(9) & 27,59(2) & 13,06(4) & 8,85(8) &  \textbf{34,19(1)} \\
      &   fault & 32,76(12) & 38,44(8) & 38,38(9) & 43,98(4) & 41,56(5) & 36,47(10) & 54,45(2) & 48,01(3) & 30,54(15) & 30,82(14) & 39,15(7) & 33,48(11) & 31,03(13) & 41,09(6) &  \textbf{63,2(1)} \\
      &    donors & 17,9(4) & 9,86(9) & 7,88(12) & 6,89(13) & 8,8(11) & 23,36(2) & 14,75(5) & 9,69(10) & 21,58(3) & 14,17(6) & 6,38(14) & 10,53(8) & 3,78(15) & 12,74(7) &  \textbf{90,85(1)} \\ 
    &     PageBlocks & 51,71(3) & 49,14(7) & 39,64(11) & 49,65(5) & 41,02(10) & 33,32(14) & 45,39(9) & 37,83(12) & 37,65(13) & 49,3(6) & 31,45(15) & 53,25(2) & 51,29(4) & 46,04(8) &  \textbf{65,26(1)} \\
    &     SpamBase & 41,57(7) & 40,12(10) & 35,16(13) & 41,18(9) & 34,73(14) & 50,03(5) & 41,42(8) & 40,03(11) &  \textbf{56,68(1)} & 53,95(3) & 42,23(6) &  NA/NA & 35,88(12) & 51,75(4) & 55,17(2) \\
      &    landsat & 16,18(15) & 16,21(14) & 24,69(7) & 30,97(2) & 24,95(6) & 22,03(10) & 24,65(8) & 26,38(4) & 17,48(13) & 25,17(5) &  \textbf{38,83(1)} & 24,48(9) & 18,86(12) & 19,81(11) & 28,89(3) \\
       &   shuttle & 92,35(7) & 85,29(8) & 13,76(14) & 60,98(9) & 12,17(15) & 96,4(4) & 20,38(11) & 20,27(12) & 96,56(2) & 95,76(5) & 15,86(13) & 93,2(6) & 48,75(10) &  \textbf{97,62(1)} & 96,52 $\pm 0,05$(3) \\  
    &     Stamps & 41,09(4) & 31,39(9) & 21,29(12) & 23,66(10) & 16,5(14) & 35,24(7) & 23,53(11) & 20,28(13) & 43,1(2) & 38,17(6) & 11,4(15) &  \textbf{43,72(1)} & 34,6(8) & 39,49(5) & 41,66(3) \\ 
      &    satellite & 59,64(7) & 57,61(9) & 37,68(15) & 61,48(6) & 39,7(14) &  \textbf{67,25(1)} & 50,01(11) & 47,23(12) & 56,58(10) & 65,94(2) & 40,11(13) & 58,33(8) & 61,94(5) & 65,92(3) & 65,73(4) \\  
      &    wine & 30,87(5) & 21,56(7) & 7,77(14) & 5,83(15) & 8,45(11) & 43,08(3) & 8,43(12) & 7,95(13) & 45,71(2) & 18,37(9) & 21,14(8) & 17,51(10) &  \textbf{48,82(1) }& 25,96(6) & 41,59(4) \\  
    &     magic.gamma & 59,27(7) & 51,43(13) & 54,76(10) & 68,85(2) & 54,12(12) & 62,41(6) & \textbf{75,63(1)} & 67,89(3) & 59,18(8) & 54,38(11) & 49,17(14) & 46,92(15) & 58,49(9) & 64,72(5) & 67,82(4) \\  
    &     campaign & 27,9(6) & 29,22(5) & 14,51(12) & 23,99(9) & 13,01(14) & 37,99(2) & 27,18(7) & 18,88(10) &  \textbf{38,58(1)} & 37,4(3) & 11,6(15) & 14,62(11) & 13,47(13) & 32,26(4) & 23,99(8) \\  
     &    http & 56,43(2) & 46,86(4) & 3,82(12) & 47,53(3) & 9,57(10) & 44,79(5) & 0,7(13) & 8,32(11) & 35,19(6) & 16,61(9) & 29,3(7) & NA/NA& 0,67(14) &  \textbf{90,83(1)} & 22,59(8) \\  
    &   InternetAds & 32,55(11) & 54,68(2) & 40,49(9) &  \textbf{58,13(1)} & 38,67(10) & 53,97(3) & 43,23(7) & 27,69(13) & 50,97(5) & 51,07(4) & 27,91(12) & NA/NA & 23,89(14) & 48,6(6) & 42,79(8) \\  
    &     fraud & 22,91(11) &  \textbf{47,58(1)} & 47,4(3) & 47,52(2) & 22,86(12) & 25,89(10) & 47,3(4) & 31,37(8) & 42,82(7) & 42,99(6) & 8,97(15) & 21,32(14) & 46,37(5) & 21,67(13) & 31,06(9) \\  
    &     census &  \textbf{10,02(1)} & 6,76(12) & 5,45(13) & 7,44(9) & 4,88(15) & 8,69(6) & 9(4) & 8,52(7) & 9,92(2) & 9,72(3) & 6,87(11) & 8,71(5) & 5,01(14) & 7,78(8) & 7,31(10) \\  
        \bottomrule
    \end{tabular}}
    
    \end{table}
\ENDGRAPH

\GRAPH ADBenchAll
\begin{table}[!ht]
\caption{AUC PR (in \%) of 15 unsupervised algorithms on 42 real-world datasets. For our method we computed the average AUC PR over 10 runs and added the standard deviation. The performance rank is shown in parenthesis (the lower, the better), and mark the best performing method(s) in
\textbf{bold}.}
\label{real_world_all}
    \centering
    \resizebox{\columnwidth}{!}{%
    \begin{tabular}{l|l|llllllllllllll|l}
    \toprule
        Category&Dataset & PCA & OCSVM & LOF & CBLOF & COF & HBOS & KNN & SOD & COPOD & ECOD & Deep SVDD & DA GMM & LODA & Iforest & \textbf{Ours} \\ \midrule
        \multirow{11}{*}{\rotatebox[origin=c]{90}{Image and CV}} & mnist & 39.93(2) & 33.2(4) & 20.9(12) & 28.82(6) & 25.51(9) & 12.51(15) & 35.53(3) & 19.15(14) & 21.35(11) & 31.93(5) & 19.72(13) & 23.75(10) & 25.86(8) & 27.71(7) & \textbf{50.17 $\pm {1.25}$ (1)} \\ 
        & optdigits & 2.76(14) & 2.92(13) & 6.06(4) & 10.08(2) & 4.42(7) & 10.03(3) & 3.06(12) & 4.39(8) & 4.36(9) & 3.43(11) & 2.5(15) & 5.59(5) & 3.95(10) & 5.09(6) &  \textbf{56.34 $\pm 5.42$(1)} \\ 
       &  skin & 17.4(12) & 19.03(7) & 18.25(10) & 29.82(2) & 16.38(13) & 23.7(6) & 28.72(3) & 24.61(5) & 17.99(11) & 15.96(14) & 18.48(8) &  NA/NA & 18.44(9) & 26.08(4) &  \textbf{52.20 $\pm 12.26$(1)}\\ 
       &  FashionMNIST & 31.42(7) & 31.97(6) & 16.85(14) & 38.9(2) & 20.73(12) & 29.43(9) & 33.87(3) & 28.72(10) & 30.32(8) & 32.53(4) & 17.43(13) & 14.44(15) & 27.32(11) & 32.35(5) &  \textbf{43.16 $\pm 2.48$(1)} \\ 
       &  MNIST-C & 16.88(8) & 17.72(7) & 13.84(13) & 27.62(2) & 14.53(12) & 15.46(11) & 22.98(3) & 15.68(10) & 15.9(9) & 18.24(5) & 8.34(15) & 11.37(14) & 18.63(4) & 17.99(6) &  \textbf{35.21 $\pm 1.26$(1)} \\ 
       &  satimage-2 & 85.69(4) & 82.71(5) & 4.29(14) &  \textbf{97.09(1)} & 8.8(13) & 78.04(7) & 39.14(10) & 26.11(11) & 76.55(8) & 63.25(9) & 3.08(15) & 22.07(12) & 80.52(6) & 93.45(3) & 94.12 $\pm 01.80$(2) \\ 
       &  MVTec-AD & 54.06(9) & 51.44(11) & 54.9(7) &  \textbf{58.52(1)} & 46.59(13) & 55.22(6) & 55.55(4) & 51.48(10) & 54.64(8) & 55.44(5) & 36.5(15) & 45.66(14) & 49.73(12) & 56.04(3) & 57.65$\pm 0.12$(2) \\ 
       &  letter & 6.86(13) & 6.1(15) &  \textbf{34.02(1)} & 14.8(6) & 21.43(5) & 8.38(10) & 30(2) & 28.63(3) & 6.77(14) & 6.94(11) & 9.29(8) & 11.68(7) & 6.87(12) & 8.49(9) & 28.47 $\pm 1.84$(4) \\ 
       &  celeba &  \textbf{15.89(1)} & 10.73(6) & 1.71(15) & 11.33(5) & 1.77(14) & 13.82(2) & 3.14(10) & 2.66(11) & 13.69(3) & 12.37(4) & 2.34(12) & 1.95(13) & 4.04(9) & 8.96(7) & 5.61 $\pm 0.50$(8) \\  
       &  CIFAR10 & 10.59(6) & 10.19(7) &  \textbf{13.02(1)} & 10.61(5) & 11.61(2) & 8.38(13) & 11.13(3) & 11.06(4) & 8.77(12) & 9.29(10) & 8.05(14) & 7.73(15) & 9.72(9) & 8.97(11) & 10.04 $\pm 0.91$ (8) \\ \midrule
\multirow{5}{*}{\rotatebox[origin=c]{90}{NLP}}     &  speech & 1.97(11) & 1.96(12) & 2.52(3) & 1.99(10) & 2.25(5) & 2.09(7) & 2.02(9) & 2.13(6) & 1.94(13) & 1.77(15) & 5.12(2) & 2.03(8) & 1.79(14) & 2.31(4) &  \textbf{7.65 $\pm 0.11$ (1)} \\  
      &   Imdb & 4.55(13) & 4.44(15) & 4.83(6) & 4.75(7) & 5.16(2) & 4.74(8) & 4.49(14) & 4.7(10) & 4.9(4) & 4.9(4) & 5.06(3) & 4.65(11) & 4.59(12) & 4.74(8) &  \textbf{7.65$\pm 0.25$ (1)} \\  

& Agnews & 5.74(9) & 5.69(10) &  \textbf{14.35(1)} & 7.02(6) & 12.21(2) & 5.58(11) & 8.61(4) & 8.4(5) & 5.43(12) & 5.43(12) & 4.45(15) & 5.41(14) & 5.93(8) & 6.04(7) & 9.04 $\pm 1.58$ (3) \\  
         & Amazon & 5.85(10) & 5.64(14) & 5.72(12) & 6.07(5) & 5.74(11) & 5.98(7) & 6.23(2) &  \textbf{6.4(1)} & 6.08(4) & 6.06(6) & 3.84(15) & 5.65(13) & 5.92(9) & 5.95(8) & 6.20 $\pm 0.23$ (3) \\  
      &   Yelp & 7.62(13) & 7.75(10) & 8.52(5) & 7.68(11) & 8.68(4) & 7.81(9) &  \textbf{9.85(1)} & 9.2(2) & 8.01(6) & 7.98(7) & 6.39(15) & 6.72(14) & 7.65(12) & 7.88(8) & 8.80 $\pm 1.32$ (3) \\ \midrule
       
\multirow{10}{*}{\rotatebox[origin=c]{90}{Healthcare}} &  WBC & 82.29(7) & 89.87(4) & 5.57(14) & 92.27(2) & 9.73(12) & 73.56(9) & 66.55(10) & 54(11) & 86.19(5) & 86.19(5) & 6.38(13) &  NA/NA & 78.67(8) & 90.49(3) &  \textbf{94.55 $\pm 4.66$(1)} \\ 
        &   Cardiotocography & 47.95(4) & 52.61(2) & 30.66(12) & 45.44(5) & 28.21(14) & 38.28(9) & 34.79(10) & 27.99(15) & 40.46(8) & 43.57(6) & 34.03(11) & 30.61(13) & 48(3) & 41.47(7) & \textbf{61.39 $\pm 5.74$(1)}\\ 
        &   Lymphography & 97.02(4) & 93.59(5) & 23.08(12) & 97.62(2) & 36.68(11) & 91.83(6) & 38.69(10) & 22.65(13) & 88.68(8) & 90.87(7) & 4.58(15) & 19.52(14) & 44.54(9) & 97.31(3) &  \textbf{99.68 $\pm 0.01$(1)} \\ 
        &   breastw & 95.11(7) & 82.7(11) & 28.55(13) & 91.54(9) & 27.6(14) & 97.71(4) & 92.19(8) & 84.88(10) &  \textbf{99.4(1) }& 98.54(2) & 50.92(12) &  NA/NA & 97.04(5) & 96.04(6) & 98.17 $\pm 0.40$(3) \\ 
        &   WPBC & 23.01(6) & 22.93(7) & 20.29(15) & 21.32(13) & 21.3(14) & 23.04(5) & 21.49(11) & 25.37(3) & 22.81(8) & 21.38(12) &  \textbf{26.24(1)} & 22.49(9) & 25.58(2) & 22.42(10) & 23.06 $\pm 0.44$(4) \\ 
        &   Hepatitis & 36.65(4) & 29.44(8) & 13.69(15) & 31.54(6) & 14.39(14) & 37.73(3) & 21.95(13) & 24.89(10) &  \textbf{41.5(1)} & 37.82(2) & 22.17(12) & 22.96(11) & 30.9(7) & 26.25(9) & 35.20 $\pm 7.17$(5) \\  
        &   thyroid & 44.34(5) & 21.23(10) & 20.81(11) & 29.95(7) & 28.5(8) & 50.98(3) & 34.98(6) & 23.56(9) & 19.64(12) & 54.05(2) & 2.5(15) & 16.06(13) & 14.68(14) &  \textbf{63.11(1)} & 47.79 $\pm 2.77$(4) \\ 
        &   annthyroid & 16.12(9) & 10.37(13) & 15.71(10) & 13.69(12) & 14.39(11) & 16.99(6) & 16.74(7) & 18.84(4) & 16.58(8) & 24.65(2) & 21.95(3) & 9.64(14) & 7.06(15) &  \textbf{30.47(1)} & 17.02 $\pm 0.80$ (5) \\  
        &   Pima & 54.03(5) & 50(8) & 47.18(10) & 53.19(6) & 44.7(11) &  \textbf{56.61(1)} & 55.14(4) & 48.24(9) & 55.19(3) & 37.3(14) & 35.87(15) & 41.55(13) & 44.09(12) & 55.82(2) & 50.09$\pm 0.89(7)$ \\ 
        &   cardio & 66.06(2) & 62.89(3) & 23.79(14) & 61.95(4) & 28.67(12) & 52.1(9) & 40.72(10) & 28.54(13) & 60.42(5) &  \textbf{68.59(1)} & 22.5(15) & 28.92(11) & 53.41(7) & 59.95(6) & 52.53 $\pm 4.68$ (8) \\  \midrule
        
\multirow{18}{*}{\rotatebox[origin=c]{90}{Others}} &  musk & 99.89(4) & 10.61(10) & 2.82(14) & 100(1) & 2.61(15) & 100(1) & 9.65(11) & 7.59(12) & 34.79(8) & 34.95(7) & 5.39(13) & 32.75(9) & 47.6(6) & 99.61(5) &  \textbf{100 $\pm 0$(1)} \\
      &    Waveform & 5.79(11) & 4.37(14) & 11.33(5) & 18.98(2) & 14.11(3) & 5.86(10) & 13.04(4) & 9.66(6) & 6.9(7) & 6.86(8) & 4.83(12) & 3.11(15) & 4.71(13) & 6.24(9) &  \textbf{33.26 $\pm 0.79$(1)} \\ 
 & cover & 9.8(7) & 11.41(5) & 8.12(9) & 5.83(13) & 4(14) & 6.83(11) & 6.16(12) & 3.88(15) & 11.37(6) & 15.63(3) & 8.12(9) & 27.59(2) & 13.06(4) & 8.85(8) &  \textbf{34.19 $\pm 0.54$(1)} \\
      &   fault & 32.76(12) & 38.44(8) & 38.38(9) & 43.98(4) & 41.56(5) & 36.47(10) & 54.45(2) & 48.01(3) & 30.54(15) & 30.82(14) & 39.15(7) & 33.48(11) & 31.03(13) & 41.09(6) &  \textbf{63.20 $\pm 1.51$ (1)} \\
      &    donors & 17.9(4) & 9.86(9) & 7.88(12) & 6.89(13) & 8.8(11) & 23.36(2) & 14.75(5) & 9.69(10) & 21.58(3) & 14.17(6) & 6.38(14) & 10.53(8) & 3.78(15) & 12.74(7) &  \textbf{90.85 $\pm 6.83$(1)} \\ 
    &     PageBlocks & 51.71(3) & 49.14(7) & 39.64(11) & 49.65(5) & 41.02(10) & 33.32(14) & 45.39(9) & 37.83(12) & 37.65(13) & 49.3(6) & 31.45(15) & 53.25(2) & 51.29(4) & 46.04(8) &  \textbf{65.26 $\pm 11.50$(1)} \\
    &     magic.gamma & 59.27(7) & 51.43(13) & 54.76(10) & 68.85(3) & 54.12(12) & 62.41(6) & 75.63(2) & 67.89(4) & 59.18(8) & 54.38(11) & 49.17(14) & 46.92(15) & 58.49(9) & 64.72(5) & \textbf{77.93 $\pm 1.29$(1)} \\  
    &     fraud & 22.91(11) &  47.58(2) & 47.4(3) & 47.52(2) & 22.86(12) & 25.89(10) & 47.3(4) & 31.37(8) & 42.82(7) & 42.99(6) & 8.97(15) & 21.32(14) & 46.37(5) & 21.67(13) & \textbf{60.08 $\pm 0.15$ (1)} \\  
    & vertebral & 10.49(10) & 10.94(8) & 14.24(3) & 11.58(6) & 13.85(4) & 9.23(14) & 10.57(9) & 11.79(5) & 8.89(15) & 11.24(7) & 10.49(10) &  15.24(2) & 9.68(13) & 10.46(12) & \textbf{20.47 $\pm 0.11$(1)} \\    

    &     SpamBase & 41.57(7) & 40.12(10) & 35.16(13) & 41.18(9) & 34.73(14) & 50.03(5) & 41.42(8) & 40.03(11) &  \textbf{56.68(1)} & 53.95(3) & 42.23(6) &  NA/NA & 35.88(12) & 51.75(4) & 55.17 $\pm 0.50$(2) \\
      &    landsat & 16.18(15) & 16.21(14) & 24.69(7) & 30.97(3) & 24.95(6) & 22.03(10) & 24.65(8) & 26.38(4) & 17.48(13) & 25.17(5) &  \textbf{38.83(1)} & 24.48(9) & 18.86(12) & 19.81(11) & 38.54 $\pm 4.54$ (2) \\
       &   shuttle & 92.35(7) & 85.29(8) & 13.76(14) & 60.98(9) & 12.17(15) & 96.4(4) & 20.38(11) & 20.27(12) & 96.56(2) & 95.76(5) & 15.86(13) & 93.2(6) & 48.75(10) &  \textbf{97.62(1)} & 96.52 $\pm 0.05$(3) \\  
    &     Stamps & 41.09(4) & 31.39(9) & 21.29(12) & 23.66(10) & 16.5(14) & 35.24(7) & 23.53(11) & 20.28(13) & 43.1(2) & 38.17(6) & 11.4(15) &  \textbf{43.72(1)} & 34.6(8) & 39.49(5) & 41.66 $\pm 0.39$(3) \\ 
      &    satellite & 59.64(7) & 57.61(9) & 37.68(15) & 61.48(6) & 39.7(14) &  67.25(2) & 50.01(11) & 47.23(12) & 56.58(10) & 65.94(3) & 40.11(13) & 58.33(8) & 61.94(5) & 65.92(4) & \textbf{73.24 $\pm 0.49$(1)} \\  
      &    wine & 30.87(5) & 21.56(7) & 7.77(14) & 5.83(15) & 8.45(11) & 43.08(3) & 8.43(12) & 7.95(13) & 45.71(2) & 18.37(9) & 21.14(8) & 17.51(10) &  \textbf{48.82(1) }& 25.96(6) & 41.59 $\pm 1.76$(4) \\  

    &     campaign & 27.9(6) & 29.22(5) & 14.51(12) & 23.99(9) & 13.01(14) & 37.99(2) & 27.18(7) & 18.88(10) &  \textbf{38.58(1)} & 37.4(3) & 11.6(15) & 14.62(11) & 13.47(13) & 32.26(4) & 23.99 $\pm 1.76$ (8) \\  
     &    http & 56.43(3) & 46.86(5) & 3.82(12) & 47.53(4) & 9.57(10) & 44.79(6) & 0.7(13) & 8.32(11) & 35.19(7) & 16.61(9) & 29.3(8) & NA/NA& 0.67(14) &  \textbf{90.83(1)} & 60.24 $\pm 0.07$(2) \\  
    &   InternetAds & 32.55(11) & 54.68(2) & 40.49(9) &  \textbf{58.13(1)} & 38.67(10) & 53.97(3) & 43.23(7) & 27.69(13) & 50.97(5) & 51.07(4) & 27.91(12) & NA/NA & 23.89(14) & 48.6(6) & 42.79 $\pm 1.43$ (8) \\  
    
    & census &  \textbf{10.02(1)} & 6.76(12) & 5.45(13) & 7.44(9) & 4.88(15) & 8.69(6) & 9(4) & 8.52(7) & 9.92(2) & 9.72(3) & 6.87(11) & 8.71(5) & 5.01(14) & 7.78(8) & 7.31 $\pm 0.10$ (10) \\  
        \bottomrule
    \end{tabular}}
    
    \end{table}
\ENDGRAPH

\subsection{Supervised anomaly detection on real world datasets}
In this subsection we used our model as fully supervised method on 29 real-world anomaly detection benchmark datasets. We then compare its performance to those of 16 state-of-the-art supervised anomaly detection algorithms. We used the same datasets are in the semi-supervised case but this time we use the true label to train our models. 
We compared the performances of our method to those of 16 baseline supervised clustering algorithms \cite{han2022adbench} including:
SVM, GANomaly \cite{akcay2019ganomaly}, DeepSAD \cite{ruff2020deep}, REPEN \cite{pang2018learning}, DevNet\cite{pang2019devNet}, PReNet\cite{pang2019deep}, FEAWAD\cite{zhou2021feature}, XGBOD\cite{zhao2018xgbod},LightGBM\cite{ke2017lightgbm}, CatBoost\cite{prokhorenkova2018catboost}, Naive Bayes (NB) \cite{bayes1763lii}, Multi-layer Perceptron (MLP) \cite{rosenblatt1958perceptron}, Random Forest (RF) \cite{breiman2001random}, XGBoost \cite{chen2016xgboost}, Residual Nets (ResNet)\cite{gorishniy2021revisiting}, FTTransformer \cite{gorishniy2021revisiting} etc.

We report the results of the supervised algorithms in table \ref{real_world_sup} and the ranking in figure \ref{fig:adbenchSup}. These results show that in the supervised scenario, Our method (AnoRand) and CatBoost have the best overall performance in terms of AUC PR and ranking. A deep analysis of the table \ref{real_world_sup} reveals that in most of the datasets, when AnoRand is not ranked first, its ranking is in the top four.
The results also show that, on most the benchmark datasets, AnoRand outperforms its counterpart deep learning based methods.
Table \ref{real_world_sup} shows that our method has better performance on big datasets such as the celeba, fraud, CIFAR10, census and the cover datasets. 
Our results also show that classical methods such as SVM, CatBoost, LGB tend to have better performance than deep learning based algorithms.

  \long\def\GRAPH #graphsgraph8 {}%

  \long\def\GRAPH #exp_resultsADBenchSup {}%

\section{Conclusion}
In this paper, we proposed a new semi supervised anomaly detection method based on deep autoencoder architecture.
This new method that we called \textbf{AnoRand}, jointly optimizes the deep autoencoder and the FFP model in an end-to-end neural network fashion. Our method is performed in two steps: we first create synthetic anomalies by randomly adding noise to few samples from the training data; secondly we train our deep learning model in supervised way with the new labeled data. 
Our method takes advantage of these limitations of FFP models in case of imbalance classes and use them to reinforce the autoencoder capabilities.
Our experimental results show that our method achieves state-of-the-art performance on synthetic datasets and 57 realworld datasets, significantly outperforming existing unsupervised alternatives. Moreover, on most the benchmark datasets whatever the category, AnoRand outperforms all its counterpart deep learning based methods.
The  We also tested our method (AnoRand) in a supervised way by using the actual labels to train the model instead of creating synthetic label as we did in the semi supervised case. The results show that it has very good performance compared to most of state-of-the-art supervised algorithms.
The main limitation of our method is that the training takes longer than most of the state-of-the-art algorithm.

\bibliographystyle{plain.bst}
\newpage
\bibliography{references}

\begin{thebibliography}{10}

\bibitem{akcay2019ganomaly}
Samet Akcay, Amir Atapour-Abarghouei, and Toby~P Breckon.
\newblock Ganomaly: Semi-supervised anomaly detection via adversarial training.
\newblock In {\em Computer Vision--ACCV 2018: 14th Asian Conference on Computer
  Vision, Perth, Australia, December 2--6, 2018, Revised Selected Papers, Part
  III 14}, pages 622--637. Springer, 2019.

\bibitem{anand1991improved}
Rangachari Anand, Kishan Mehrotra, Chilukuri~K Mohan, and Sanjay Ranka.
\newblock An improved algorithm for neural network classification of imbalanced
  training sets.
\newblock 1991.

\bibitem{bayes1763lii}
Thomas Bayes.
\newblock Lii. an essay towards solving a problem in the doctrine of chances.
  by the late rev. mr. bayes, frs communicated by mr. price, in a letter to
  john canton, amfr s.
\newblock {\em Philosophical transactions of the Royal Society of London},
  (53):370--418, 1763.

\bibitem{breiman2001random}
Leo Breiman.
\newblock Random forests.
\newblock {\em Machine learning}, 45:5--32, 2001.

\bibitem{breunig2000lof}
Markus~M Breunig, Hans-Peter Kriegel, Raymond~T Ng, and J{\"o}rg Sander.
\newblock Lof: identifying density-based local outliers.
\newblock In {\em Proceedings of the 2000 ACM SIGMOD international conference
  on Management of data}, pages 93--104, 2000.

\bibitem{cai2022perturbation}
Jinyu Cai and Jicong Fan.
\newblock Perturbation learning based anomaly detection.
\newblock In Alice~H. Oh, Alekh Agarwal, Danielle Belgrave, and Kyunghyun Cho,
  editors, {\em Advances in Neural Information Processing Systems}, 2022.

\bibitem{chawla2002smote}
Nitesh~V Chawla, Kevin~W Bowyer, Lawrence~O Hall, and W~Philip Kegelmeyer.
\newblock Smote: synthetic minority over-sampling technique.
\newblock {\em Journal of artificial intelligence research}, 16:321--357, 2002.

\bibitem{chen2016xgboost}
Tianqi Chen and Carlos Guestrin.
\newblock Xgboost: A scalable tree boosting system.
\newblock In {\em Proceedings of the 22nd acm sigkdd international conference
  on knowledge discovery and data mining}, pages 785--794, 2016.

\bibitem{collin2021improved}
Anne-Sophie Collin and Christophe De~Vleeschouwer.
\newblock Improved anomaly detection by training an autoencoder with skip
  connections on images corrupted with stain-shaped noise.
\newblock In {\em 2020 25th International Conference on Pattern Recognition
  (ICPR)}, pages 7915--7922. IEEE, 2021.

\bibitem{cortes1995support}
Corinna Cortes and Vladimir Vapnik.
\newblock Support-vector networks.
\newblock {\em Machine learning}, 20:273--297, 1995.

\bibitem{Devlin2019BERTPO}
Jacob Devlin, Ming-Wei Chang, Kenton Lee, and Kristina Toutanova.
\newblock Bert: Pre-training of deep bidirectional transformers for language
  understanding.
\newblock {\em ArXiv}, abs/1810.04805, 2019.

\bibitem{goldstein2012histogram}
Markus Goldstein and Andreas Dengel.
\newblock Histogram-based outlier score (hbos): A fast unsupervised anomaly
  detection algorithm.
\newblock {\em KI-2012: poster and demo track}, 1:59--63, 2012.

\bibitem{gorishniy2021revisiting}
Yury Gorishniy, Ivan Rubachev, Valentin Khrulkov, and Artem Babenko.
\newblock Revisiting deep learning models for tabular data.
\newblock {\em Advances in Neural Information Processing Systems},
  34:18932--18943, 2021.

\bibitem{han2022adbench}
Songqiao Han, Xiyang Hu, Hailiang Huang, Minqi Jiang, and Yue Zhao.
\newblock Adbench: Anomaly detection benchmark.
\newblock {\em Advances in Neural Information Processing Systems},
  35:32142--32159, 2022.

\bibitem{he2016deep}
Kaiming He, Xiangyu Zhang, Shaoqing Ren, and Jian Sun.
\newblock Deep residual learning for image recognition.
\newblock In {\em Proceedings of the IEEE conference on computer vision and
  pattern recognition}, pages 770--778, 2016.

\bibitem{he2003discovering}
Zengyou He, Xiaofei Xu, and Shengchun Deng.
\newblock Discovering cluster-based local outliers.
\newblock {\em Pattern recognition letters}, 24(9-10):1641--1650, 2003.

\bibitem{ke2017lightgbm}
Guolin Ke, Qi~Meng, Thomas Finley, Taifeng Wang, Wei Chen, Weidong Ma, Qiwei
  Ye, and Tie-Yan Liu.
\newblock Lightgbm: A highly efficient gradient boosting decision tree.
\newblock {\em Advances in neural information processing systems}, 30, 2017.

\bibitem{kingma2013auto}
Diederik~P Kingma and Max Welling.
\newblock Auto-encoding variational bayes.
\newblock {\em arXiv preprint arXiv:1312.6114}, 2013.

\bibitem{kriegel2009outlier}
Hans-Peter Kriegel, Peer Kr{\"o}ger, Erich Schubert, and Arthur Zimek.
\newblock Outlier detection in axis-parallel subspaces of high dimensional
  data.
\newblock In {\em Advances in Knowledge Discovery and Data Mining: 13th
  Pacific-Asia Conference, PAKDD 2009 Bangkok, Thailand, April 27-30, 2009
  Proceedings 13}, pages 831--838. Springer, 2009.

\bibitem{li2020copod}
Zheng Li, Yue Zhao, Nicola Botta, Cezar Ionescu, and Xiyang Hu.
\newblock Copod: copula-based outlier detection.
\newblock In {\em 2020 IEEE international conference on data mining (ICDM)},
  pages 1118--1123. IEEE, 2020.

\bibitem{li2022ecod}
Zheng Li, Yue Zhao, Xiyang Hu, Nicola Botta, Cezar Ionescu, and George Chen.
\newblock Ecod: Unsupervised outlier detection using empirical cumulative
  distribution functions.
\newblock {\em IEEE Transactions on Knowledge and Data Engineering}, 2022.

\bibitem{4781136}
Fei~Tony Liu, Kai~Ming Ting, and Zhi-Hua Zhou.
\newblock Isolation forest.
\newblock In {\em 2008 Eighth IEEE International Conference on Data Mining},
  pages 413--422, 2008.

\bibitem{pang2018learning}
Guansong Pang, Longbing Cao, Ling Chen, and Huan Liu.
\newblock Learning representations of ultrahigh-dimensional data for random
  distance-based outlier detection.
\newblock In {\em Proceedings of the 24th ACM SIGKDD international conference
  on knowledge discovery \& data mining}, pages 2041--2050, 2018.

\bibitem{pang2019deep}
Guansong Pang, Chunhua Shen, Huidong Jin, and Anton van~den Hengel.
\newblock Deep weakly-supervised anomaly detection.
\newblock {\em arXiv preprint arXiv:1910.13601}, 2019.

\bibitem{pang2019devNet}
Guansong Pang, Chunhua Shen, and Anton van~den Hengel.
\newblock Deep anomaly detection with deviation networks.
\newblock In {\em Proceedings of the 25th ACM SIGKDD international conference
  on knowledge discovery \& data mining}, pages 353--362, 2019.

\bibitem{pevny2016loda}
Tom{\'a}{\v{s}} Pevn{\`y}.
\newblock Loda: Lightweight on-line detector of anomalies.
\newblock {\em Machine Learning}, 102:275--304, 2016.

\bibitem{prokhorenkova2018catboost}
Liudmila Prokhorenkova, Gleb Gusev, Aleksandr Vorobev, Anna~Veronika Dorogush,
  and Andrey Gulin.
\newblock Catboost: unbiased boosting with categorical features.
\newblock {\em Advances in neural information processing systems}, 31, 2018.

\bibitem{ramaswamy2000efficient}
Sridhar Ramaswamy, Rajeev Rastogi, and Kyuseok Shim.
\newblock Efficient algorithms for mining outliers from large data sets.
\newblock In {\em Proceedings of the 2000 ACM SIGMOD international conference
  on Management of data}, pages 427--438, 2000.

\bibitem{rosenblatt1958perceptron}
Frank Rosenblatt.
\newblock The perceptron: a probabilistic model for information storage and
  organization in the brain.
\newblock {\em Psychological review}, 65(6):386, 1958.

\bibitem{ruff2018deep}
Lukas Ruff, Robert Vandermeulen, Nico Goernitz, Lucas Deecke, Shoaib~Ahmed
  Siddiqui, Alexander Binder, Emmanuel M{\"u}ller, and Marius Kloft.
\newblock Deep one-class classification.
\newblock In {\em International conference on machine learning}, pages
  4393--4402. PMLR, 2018.

\bibitem{ruff2020deep}
Lukas Ruff, Robert~A. Vandermeulen, Nico G{\"o}rnitz, Alexander Binder,
  Emmanuel M{\"u}ller, Klaus-Robert M{\"u}ller, and Marius Kloft.
\newblock Deep semi-supervised anomaly detection.
\newblock In {\em International Conference on Learning Representations}, 2020.

\bibitem{saito2015precision}
Takaya Saito and Marc Rehmsmeier.
\newblock The precision-recall plot is more informative than the roc plot when
  evaluating binary classifiers on imbalanced datasets.
\newblock {\em PloS one}, 10(3):e0118432, 2015.

\bibitem{scholkopf1999support}
Bernhard Sch{\"o}lkopf, Robert~C Williamson, Alex Smola, John Shawe-Taylor, and
  John Platt.
\newblock Support vector method for novelty detection.
\newblock {\em Advances in neural information processing systems}, 12, 1999.

\bibitem{shi2021unsupervised}
Yong Shi, Jie Yang, and Zhiquan Qi.
\newblock Unsupervised anomaly segmentation via deep feature reconstruction.
\newblock {\em Neurocomputing}, 424:9--22, 2021.

\bibitem{shyu2003novel}
Mei-Ling Shyu, Shu-Ching Chen, Kanoksri Sarinnapakorn, and LiWu Chang.
\newblock A novel anomaly detection scheme based on principal component
  classifier.
\newblock Technical report, Miami Univ Coral Gables Fl Dept of Electrical and
  Computer Engineering, 2003.

\bibitem{tang2002enhancing}
Jian Tang, Zhixiang Chen, Ada Wai-Chee Fu, and David~W Cheung.
\newblock Enhancing effectiveness of outlier detections for low density
  patterns.
\newblock In {\em Advances in Knowledge Discovery and Data Mining: 6th
  Pacific-Asia Conference, PAKDD 2002 Taipei, Taiwan, May 6--8, 2002
  Proceedings 6}, pages 535--548. Springer, 2002.

\bibitem{you2022a}
Zhiyuan You, Lei Cui, Yujun Shen, Kai Yang, Xin Lu, Yu~Zheng, and Xinyi Le.
\newblock A unified model for multi-class anomaly detection.
\newblock In Alice~H. Oh, Alekh Agarwal, Danielle Belgrave, and Kyunghyun Cho,
  editors, {\em Advances in Neural Information Processing Systems}, 2022.

\bibitem{zavrtanik2021draem}
Vitjan Zavrtanik, Matej Kristan, and Danijel Sko{\v{c}}aj.
\newblock Draem-a discriminatively trained reconstruction embedding for surface
  anomaly detection.
\newblock In {\em Proceedings of the IEEE/CVF International Conference on
  Computer Vision}, pages 8330--8339, 2021.

\bibitem{zhao2018xgbod}
Yue Zhao and Maciej~K Hryniewicki.
\newblock Xgbod: improving supervised outlier detection with unsupervised
  representation learning.
\newblock In {\em 2018 International Joint Conference on Neural Networks
  (IJCNN)}, pages 1--8. IEEE, 2018.

\bibitem{zhao2019pyod}
Yue Zhao, Zain Nasrullah, and Zheng Li.
\newblock Pyod: A python toolbox for scalable outlier detection.
\newblock {\em Journal of Machine Learning Research}, 20(96):1--7, 2019.

\bibitem{zhou2020encoding}
Kang Zhou, Yuting Xiao, Jianlong Yang, Jun Cheng, Wen Liu, Weixin Luo, Zaiwang
  Gu, Jiang Liu, and Shenghua Gao.
\newblock Encoding structure-texture relation with p-net for anomaly detection
  in retinal images.
\newblock In {\em Computer Vision--ECCV 2020: 16th European Conference,
  Glasgow, UK, August 23--28, 2020, Proceedings, Part XX 16}, pages 360--377.
  Springer, 2020.

\bibitem{zhou2021feature}
Yingjie Zhou, Xucheng Song, Yanru Zhang, Fanxing Liu, Ce~Zhu, and Lingqiao Liu.
\newblock Feature encoding with autoencoders for weakly supervised anomaly
  detection.
\newblock {\em IEEE Transactions on Neural Networks and Learning Systems},
  33(6):2454--2465, 2021.

\bibitem{zong2018deep}
Bo~Zong, Qi~Song, Martin~Renqiang Min, Wei Cheng, Cristian Lumezanu, Daeki Cho,
  and Haifeng Chen.
\newblock Deep autoencoding gaussian mixture model for unsupervised anomaly
  detection.
\newblock In {\em International conference on learning representations}, 2018.

\end{thebibliography}


\end{document}